\documentclass[conference]{IEEEtran}
\usepackage{booktabs}
\usepackage{cite}
\usepackage{textcomp}
\usepackage{graphicx}
\usepackage{blindtext}
\usepackage{soul}
\usepackage[table]{xcolor}
\usepackage{hyperref}
\usepackage{amsmath,amsfonts,amsthm,bm} 
\usepackage{multirow}
\usepackage{enumerate}
\usepackage{subcaption}
\usepackage[linesnumbered,ruled]{algorithm2e}
\usepackage{pifont}
\usepackage[acronym]{glossaries}
\usepackage{rotating}
\usepackage{float}
\usepackage{array}
\usepackage{ctable}
\usepackage{txfonts}

\def\BibTeX{{\rm B\kern-.05em{\sc i\kern-.025em b}\kern-.08em
    T\kern-.1667em\lower.7ex\hbox{E}\kern-.125emX}}

\hypersetup{colorlinks,  urlcolor=blue, hidelinks} 

\newcommand{\comment}[1]{ }

\newcommand{\rpm}{\raisebox{.2ex}{$\scriptstyle\pm$}}

\newcommand{\cmark}{\text{\ding{51}}}
\newcommand{\xmark}{\text{\ding{55}}}

\newcommand{\tr}{train}
\newcommand{\tst}{test}

\newcommand{\IEEEcopyright}{\copyright 2021 IEEE. Personal use of this material is permitted. Permission
from IEEE must be obtained for all other uses, in any current or future
media, including reprinting/republishing this material for advertising or
promotional purposes, creating new collective works, for resale or
redistribution to servers or lists, or reuse of any copyrighted
component of this work in other works.}

\newcommand\blfootnote[1]{%
  \begingroup
  \renewcommand\thefootnote{}\footnote{#1}%
  \addtocounter{footnote}{-1}%
  \endgroup
}

\renewcommand{\arraystretch}{0.4}

\SetCommentSty{mycommfont}
\SetKwInput{KwInput}{Require}                
\SetKwInput{KwOutput}{Output}              

\graphicspath{ {./figures/} }

\newacronym{iks}{IKS}{Incremental Kolmogorov–Smirnov test}
\newacronym{hdddm}{HDDDM}{Hellinger Distance Drift Detector Method}
\newacronym{zscore}{ZSD}{Z-Score Detector}
\newacronym{ema}{EMAD}{Exponential Moving Average Detector}
    
\begin{document}

\title{Task-Sensitive Concept Drift Detector with Constraint Embedding}

\author{\IEEEauthorblockN{Andrea Castellani}
\IEEEauthorblockA{\textit{CITEC, Bielefeld University} \\
Bielefeld, Germany \\
acastellani@techfak.uni-bielefeld.de}
\and
\IEEEauthorblockN{Sebastian Schmitt}
\IEEEauthorblockA{\textit{Honda Research Institute Europe GmbH}\\
Offenbach, Germany \\
sebastian.schmitt@honda-ri.de}
\and
\IEEEauthorblockN{Barbara Hammer}
\IEEEauthorblockA{\textit{CITEC, Bielefeld University} \\
Bielefeld, Germany \\
bhammer@techfak.uni-bielefeld.de}
}

\maketitle

\begin{abstract}
Detecting drifts in data is essential for machine learning applications,  as changes in the statistics of processed data typically has a profound influence on the performance of trained  models. 
Most of the available drift detection methods are either supervised and require access to the true labels during inference time, or they are completely unsupervised and aim for changes in distributions without taking label information into account. 
We propose a novel task-sensitive semi-supervised drift detection scheme, which utilizes label information while training the initial model, but takes into account that supervised label information is no longer available when using the model during inference.
It utilizes a constrained low-dimensional embedding representation of the input data. This way, it is best suited for the classification task.
It is able to detect real drift, where the drift affects the classification performance, while it properly ignores virtual drift, where the classification performance is not affected by the drift.   
In the proposed framework, the actual method to detect a change in the statistics of incoming data samples can be chosen freely.
Experimental evaluation on nine benchmarks datasets, with different types of drift, demonstrates that the proposed framework can reliably detect drifts, and outperforms state-of-the-art unsupervised drift detection approaches.

\end{abstract}


\section{Introduction}
\blfootnote{\IEEEcopyright}
In the context of data-driven machine learning, and data mining applications, the dataset used for calibrating or training the models plays a central role. 
Its statistical properties define the model's behavior and introduce implicit or explicit assumptions. 
In many situations, it is assumed that the distribution of the data streams is stationary, i.e.\ not changing over time, which is a valid assumption if the distribution of the data used to calibrate/train the model is the same as later during production use.
However, due to various reasons, such as aging or slight system re-configurations, this assumption is not met in many real-world applications \cite{Gama2014ASO}.
Usually, a drift has a strong impact on model performance if it occurs unexpectedly.

In literature, this phenomenon is usually referred to as \textit{concept drift} \cite{Gama2014ASO} and it has received a lot of attention in the past few years \cite{Gemaque2020AnOO,Hu2020NoFL,Lu2019LearningUC}.
Concept drift is defined as the change in the joint distribution of a set of input variables $\bm{x}$ and target variable $\bm{y}$ over time, i.e.\ $P_{t_0}(\bm{x}, \bm{y}) \neq P_{t_1}(\bm{x}, \bm{y})$ where $t_0$ and $t_1$ are usually \textit{training} and \textit{testing} time.
We focus on classification tasks, where the output variables $\bm{y}$ represent class labels.
Two fundamentally different categories exist for concept drift. \textit{Virtual drift} \cite{Gama2014ASO} refers to changes in the distribution of the input data $\bm{x}$, without affecting the distribution of the classification labels, i.e.\  $P_{\tr}(\bm{x}) \neq P_{\tst}(\bm{x})$ but  $P_{\tr}(\bm{y}|\bm{x}) = P_{\tst}(\bm{y}|\bm{x})$.
\textit{Real concept drift} refers to any change in the distribution $P(\bm{y}|\bm{x})$ which affects the classification labels.
Another crucial aspect for concept drift is given by the speed of the changes. A \textit{gradual concept drift} characterizes the case where the transition occurs smoothly over time, while for a \textit{step concept drift}  the switch between two contexts occurs abruptly.

Concept drift detection is an important task for many real-world classification tasks, such as anomaly detection, fraud detection or monitoring the electricity load profiles of an industrial facility \cite{ZliobaiteOverview2016, castellaniDTwin21, castellani2021estimating}, as it can provide additional information to improve the performance. 
Interestingly, the majority of concept drift detectors operate in a \textit{supervised} fashion and  require immediate access to ground-truth labels during classifier training \textit{and} testing/inference.
But for many of the aforementioned applications, the acquisition of true labels is intractable or at least very expensive. 
Therefore, labels are possibly only available for training and testing where they can be utilized for building the classifier model, but they are typically not available during inference in the real application scenario. 
These type of situations are addressed by \textit{unsupervised} or \textit{semi-supervised} drift detection methods, which is a much less explored research area \cite{Gemaque2020AnOO}. 
Moreover, there exist fully unsupervised drift detection methods which aim for anomalies of an underlying data distribution, but do not take into account labels available during training \cite{Reis2016FastUO}.

\begin{figure}[tb]
    \centering
    \includegraphics[width=0.9\linewidth]{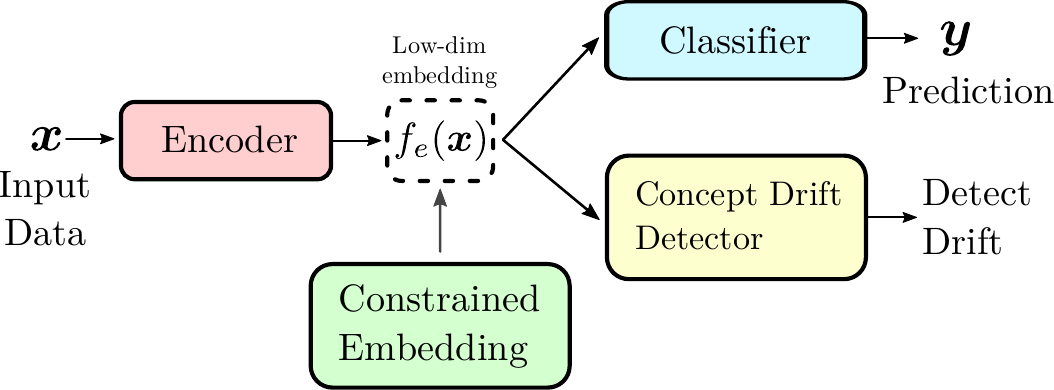}
    \caption{Proposed Concept Drift architecture.}
    \label{fig:proposed}
\end{figure}

In this work, we propose a  task-sensitive concept drift detection framework, where classification labels are only necessary during training of the classifier. 
We  learn a meaningful low-dimensional embedding representation of the input data, which is best suited for the classification task. 
Additionally, we employ a constraint in that embedding representation, which forces the latent representation to have small intra-class and large inter-class distances, and hence it optimally represents the given class structure.
After training, concept drifts can be detected by monitoring the distances of incoming new data samples to the learned class centroids in the embedding space. This is highly sensitive to the trained classification task, but does not need true class labels during inference. The actual detection of the drift can be performed by analyzing the distance statistics derived from the embedding representation with any state-of-the-art (SotA) drift detection method. 
The proposed concept drift architecture is reported in Fig.~\ref{fig:proposed}.

In addition, 
we provide a number of minor improvements: (i) two simple drift detection methods based on the exponential moving average (EMAD), and a modified $z$-score (ZSD); (ii) a novel evaluation metric ($H$-score), which accumulates three SotA drift detection metrics 
into one performance measure; (iii) an extensive evaluation of the proposed framework including ablation studies on the introduced hyperparameters on various datasets for different types of induced concept drift.

\section{Related work}\label{sec:related}

The studies on the concept drift can be divided in two groups \cite{Hu2020NoFL}: \textit{performance-based} and \textit{data distribution-based} approaches.
Performance-based techniques aims at tracking changes in the error rate of a model. 
These techniques require access to the ground-truth labels.
Popular algorithms are {Drift Detection Model} (DDM) \cite{Gama2004LearningWD} and {Adaptive Windowing} (ADWIN) \cite{Bifet2007LearningFT}.
Data-distribution based approaches monitor distribution changes in the data, and are based on statistical tests for distribution similarity.
An example are {Hellinger Distance Drift Detection Model} (HDDDM) \cite{Ditzler2011HellingerDB} and {Incremental Kolomogorow-Smirnov test} (IKS) \cite{Reis2016FastUO}.
Those algorithms can work with unlabeled data, 
but they are quite sensitive to data changes and, considering solely input data, bears the risk of detecting drifts in features that may be not important for the prediction model, i.e.\ wrongly detect virtual drifts.
The proposed framework targets to only detect real drifts, i.e.\ changes in the input data that also have an impact on the classification results.

There have been few tentative attempts to develop unsupervised concept drift detection methods by using neural networks.
In \cite{Jaworski2020ConceptDD}, the authors use the reconstruction error of an autoencoder to detect changes in the data. 
In \cite{Yang2020CADEDA}, they use a contrastive loss to learn a low-dimensional embedding of data.
Both of those  methods, are sensitive to virtual drifts because they rely solely on raw input data and ignore the classification task.
The internal representation learned by neural network is often utilized also for other purposes, rather than only fulfilling the main classification task. 
In \cite{Li2019SemisupervisedCW} clustering is applied to the learned representation of the data, rather than in the original space.
The authors of \cite{Dube2020AutomatedDO} and \cite{Ackerman2020SequentialDD} propose to detect drift by using the embedding of a neural network classifier. 
This method can correctly ignore virtual drifts, but it needs to compute a reference statistic on the training data, which could be difficult to compute while working with large datasets.
In contrast, we use very simple distance statistics in the embedding representation and derive the reference statistics from a small data sample with an additional simple update rule to detect drifted samples.

\section{Proposed Framework}\label{sec:proposed}

\subsection{Constrained Embedding Representation}
We assume that the dataset consists of $n$ data samples $(\bm{x}_i, \bm{y}_i)$ with $i=1,\dots,n$, where $\bm{x}_i \in \mathbb{R}^{q}$ represents  $q$-dimensional raw input data and $\bm{y}_i \in \{0, 1\}^k$ the corresponding ground truth label, indicating to which of the $k$-classes the sample belongs (one-hot encoding).

We train a neural network that consists of an \textit{encoder} part $f_e$, which transforms the input data $\bm{x}_i$ into a low-dimensional embedding representation, $\bm{e}_i=f_e(\bm{x}_i) \in \mathbb{R}^d$ with $d\ll q$. A \textit{classifier} neural network $f_c$ takes this low dimensional embedding and produces a prediction of the class labels as output, 
$\bm{p}_i = \text{softmax}\big(f_c(\bm{e}_i)\big)$.
The loss function for the classification task is then given by the cross entropy,
\begin{equation}\label{eq:CE}
    \mathcal{L}_{c} = -\frac{1}{n} \sum_{i=1}^{n}  \bm{y}_{i}^T\cdot \log (\bm{p}_{i}).
\end{equation}

Additionally, we include a \textit{constrained embedding module} $f_{ce}$, \cite{castellani2021estimating}  which forces the latent representation to have small intra-class and large inter-class distances.
For this, we first  initialize the $k$ centroids $\bm{C}_j \in \mathbb{R}^{d}$ with $j=1,\dots,k$, in the \textit{d}-dimensional embedding space randomly.
Then, we iteratively adapt $\bm{C}_j$ in order to minimize the intra-class and maximize the inter-class distances. 
The loss function is given by \cite{castellani2021estimating}:
\begin{multline}\label{eq:CC}
    \mathcal{L}_{ce} = \frac{1}{n}\sum_{i=1}^{n} \Bigg[ {\underbrace{  \left \| f_e(\bm{x}_i) - \bm{C}_{\bm{y}_i} \right \| ^ 2 _2}_{\text{intra-class}}} + \\
    + {\underbrace{\log \sum_{j=1}^{k} \exp \left ( - \left \| f_e(\bm{x}_i) - \bm{C}_j \right \| _2 \right ) }_{\text{inter-class}}} \Bigg]  + \ell_{reg},
\end{multline}
where $\bm{C}_{\bm{y}_i}$ indicates the centroid representing the ground truth class label of sample $i$.    
The regularization term $\ell_{reg} = - \sum_l^k \min_{l \neq j} \log \left \| \bm{C}_l- \bm{C}_j \right \|_2$ aims to create well separated embedding for different classes.

The final total loss function is given by the sum of those contributions: $\mathcal{L} = \mathcal{L}_c + \mathcal{L}_{ce}$.
During training, all parameters of the neural networks, as well as the constraint clustering, are learned and the procedure is summarized in Algorithm~\ref{alg:constrained_training}.

\subsection{Task-Sensitive Concept Drift Detection Framework}
In order to detect drifting samples during inference, without access to ground-truth labels, we take the fully trained classifier $f_c$ with the constrained embedding $f_{ce}$, described in the previous section, and analyze the data statistics in the embedding space. 
As the constrained embedding representation is trained with the classifier, we expect that the drift detector is able to detect drifts only if it is affecting the classification task, i.e.\ real  task-sensitive drift. 
Since the embedding space is tailored toward a compact representation of classes, we rely on particularly simple distance-based statistics to describe samples within the embedding space.

Let $\bm{x}_i \in \bm{X}^{test}$ be a sample of a data stream, the idea is to calculate the Euclidean distances of each sample to the class centroids $\bm{C}_j$ in the latent space:
\begin{equation} \label{eq:pdist}
    \bm{d}_{i,j} = \| f_e(\bm{x}_i) - \bm{C}_j \|_2 \,.
\end{equation} 

The centroids $\bm{C}_j$ are fixed after the training and the distances are expected to be small w.r.t.\ the centroid of the predicted class, and large to the  centroids representing the other classes. Therefore, we extract different statistical feature vectors $\bm{m}_{i}$ from the distance statistics of sample $i$ w.r.t.\ all centroids, $\{\bm{d}_{i,j}\}_{j=1}^{k}$, and use them  for drift detection.

The actual drift detection is done by comparing the feature statistics of each sample to a reference distribution, $\{\bm{m}^{ref}\}$, which is calculated from some small reference data set and which thereby defines the non-drifted data. 
Then, the statistics $\bm{m}_i$ for the sample $i$, is compared against $\{\bm{m}^{ref}\}$ by a change detection method $\mathcal{DM}$:
\begin{equation}
    o_i = \mathcal{DM}(\bm{m}_i, \{\bm{m}^{ref}\}),
\end{equation}
where $o_i \in [0, 1]$ is a binary flag that signals if the test sample is significantly different from the reference distribution.
Note that $\mathcal{DM}$ can be any unsupervised change detection method.

Finally, in order to reduce the ratio of false detections, we declare that a drift happened only if the number of raw detections $o_i$ in the last $w$ samples is higher than a detection threshold $r$. Precisely, we report a drift at the time  $i$ if
\begin{equation}\label{eq:drift_detection}
    \frac1{w} \sum_{t=0}^{w-1} o_{i-t} > r.
\end{equation}
In case no drift is detected, the reference statistics $\{\bm{m}^{ref}\}$ is updated with the test sample statistic $\bm{m}_i$ of undrifted samples, i.e.\ when  $o_i=0$.
The pseudo-code of the proposed concept drift detection framework is reported in Algorithm~\ref{alg:drift_detection}.

\begin{algorithm}[t]
\DontPrintSemicolon
\footnotesize
  \KwInput{Data $\{ (\bm{x}_i, \bm{y}_i)\}_n$, training model $\mathcal{M}$: classifier $f_c$, encoder $f_e$, constraint module $f_{ce}$}
  \KwOutput{Trained model $\mathcal{M}$: $f_c$, $f_e$, $f_{ce}$, centroids $\{\bm{C}_j\}_k$}
  
  $\{\bm{C}_j\} \gets $ random initialization \tcp*{Initialize the centorids} 
  
  \For{\textit{training epoch} $t = 0$ \KwTo $t_{end}$}{
    Fetch mini-batch data $\{ (\bm{x}_i, \bm{y}_i)\}_b$ at current epoch $t$\\
    $\bm{e}_i = f_e (\bm{x}_i)$ \tcp*{Embedding forward pass}
    $\bm{p}_i = \text{softmax}(f_c(\bm{e}_i))$ \tcp*{Classifer forward pass}
    $\mathcal{L}_{c}(\bm{p}_i, \bm{y}_i) \gets $ Eq. \ref{eq:CE} \tcp*{Classification loss}
    $\mathcal{L}_{ce}(\bm{e}_i, \{\bm{C}_j\}) \gets $ Eq. \ref{eq:CC} \tcp*{Constrained embedding loss}
    Update $\mathcal{M}$ by SGD on $\mathcal{L}_c + \mathcal{L}_{ce}$ \tcp*{Backward pass}
}
\caption{Training with constrained embedding}\label{alg:constrained_training}
\end{algorithm}

\begin{algorithm}[t]
\DontPrintSemicolon
\footnotesize
\KwInput{Trained encoder $f_e$, centroids $\{\bm{C}_j\}_k$, reference data $\{\bm{X}\}^{ref}$, test data $\{\bm{X}\}^{test}$, unsupervised detector $\mathcal{DM}$, detection history size $w$, detection threshold $r$}
\KwOutput{Drift detection index $i_d$}

$i = 0$ , $i_d = \infty$ \tcp*{Init sample and drift index}

\tcc{Calculate reference statistics}
\For{$\bm{x}_r$ in $\{ \bm{X}^{ref} \}$} {
$\bm{d}_{r,j}= \| f_e (\bm{x}_r) - \bm{C}_j \|_2$ \tcp*{Distances to centroids} 
$\{\bm{m}^{ref}\} \gets$ get statistical features of $\{\bm{d}_{r,j}\}_{j=1}^{k}$ \tcp*{Reference statistics}
}

\tcc{Begin Concept Drift detection}
\For{$\bm{x}_i$ in $\{\bm{X}^{test}\}$}{
$\bm{d}_{i,j} = \| f_e (\bm{x}_i) - \bm{C}_j \|_2$ \tcp*{Distances to centroids} 
$\bm{m}_i \gets $ get statistical features of $\{\bm{d}_{i,j}\}_{j=1}^{k}$ \tcp*{Get statistics}

$o_i = \mathcal{DM}(\bm{m}_i, \{\bm{m}^{ref}\})$ \tcp*{Compare statistics}

\uIf{$\frac1{w} \sum_{t=0}^{w-1} o_{i-t} > r $}{
$i_d = i$  \tcp*{Drift detected, end algorithm}
break\;
}
\ElseIf{$o_i=0$}{
$\{\bm{m}^{ref}\} \gets \bm{m}_i$ \tcp*{Update reference statistics}
}
$i=i+1$\;
}

\caption{Task-sensitive drift detection framework}\label{alg:drift_detection}
\end{algorithm}

\subsection{Unsupervised Drift Detection Methods}
The proposed drift detection framework is flexible to be used with any unsupervised change detection algorithm.
We propose two novel detectors, which are tailored to our specific representation and are particularly simple. 
They are efficient and allow for continuous self-calibration of their meta-parameters. 

\subsubsection{\textbf{\acrfull{ema}}}
It monitors the exponentially weighted running statistics of the distance of each sample to its closest centroid.
In this case the feature vector $m_{i}$  just has one entry and is defined as 
\begin{equation}\label{eq:dmin}
    \displaystyle m_i = \min_{j} \bm{d}_{i,j}\,.
\end{equation}
For the change detector we calculate the  exponential moving averages of the feature and its variance,  
\begin{align}
    \mu_i & = \lambda  \cdot \mu_{i-1} + (1 - \lambda) \cdot  m_{i} \label{eq:ema} \\
    \sigma_i & = \lambda \cdot \sigma_{i-1} + (1 - \lambda) \cdot (m_{i} - \mu_i)^2 \label{eq:emsd} 
\end{align}
with the forgetting factor $\lambda = 0.95$ and $i$ the time stamp.

We detect a change when the actual average $\mu_i$ is greater than a threshold $\beta$, i.e.\ $\mu_i > \beta$.
Here, $\beta$ is an adaptive threshold which is  initialized with the reference-set $\bm{X}^{ref}$ statistic, 
\begin{equation}
    \beta = \mu_{\bm{m}_r} + \sigma_{\bm{m}_r}.
\end{equation}
Note that we use $\mu + \sigma$ as threshold as we only flag when the actual metric is \textit{significantly greater} than the reference (not when it is smaller).
When no change is detected, the threshold $\beta$ is updated with the current statistics: $\beta = \mu_i + \sigma_i$.

\subsubsection{\textbf{\acrfull{zscore}}}
The feature is given by the distance of each sample to its closest centroid $m_{i}$, Eq.~\eqref{eq:dmin}. 
The feature is transformed to a $z$-score $z$, using the mean ($\mu$) and standard deviation ($\sigma$) of the reference statistic $\bm{m}_r$,
\begin{equation}
    z_i = \frac{m_i - \mu_{\bm{m}_r}}{\sigma_{\bm{m}_r}}
\end{equation}
We then compute a $p$-value using the cumulative distribution function of the  normal distribution $\Phi$ as $p_i = \Phi(z_i)$.
A change is detected if the $p$-value is below a confidence level $\alpha$, i.e.\  $p_i < \alpha$ with $\alpha=0.05$.
When no change is detected, the $m_{i}$ is used to update the reference statistic $\bm{m}_r$, and  the $\mu_{\bm{m}_r}$ and $\sigma_{\bm{m}_r}$ are dynamically updated with \eqref{eq:ema} and \eqref{eq:emsd} respectively.

\section{Experimental setup}

\subsection{Comparative methods}
We also report results using state-of-the-art unsupervised detection methods which are briefly describe in the following. 

\subsubsection{\textbf{\acrfull{hdddm}}}
It monitors the Hellinger distance to detect a drift between two multivariate distributions \cite{Ditzler2011HellingerDB}.
Since \acrshort{hdddm} is a batch-based method, we modified this approach in the same way as \cite{Sethi2017OnTR}, such that a batch is defined incrementally by a sliding window of $5 w$ samples, with a stride of one sample.
Additionally, to detect a change, we directly use the Hellinger distance $\delta_H$:  $\delta_H > \beta$, with $\beta=\mu + \sigma$, as in \cite{Ditzler2011HellingerDB}.
The threshold is initialized with the first $5 w$ samples to calculate the statistics $\mu$ and $\sigma$. 

We do experiment with \acrshort{hdddm} on both, the raw input data $\bm{x}$, and on the network low-dimensional embedding $f_e(\bm{x})$. 
In the remainder of the paper, the methods are referred as \acrshort{hdddm}$_I$ and \acrshort{hdddm}$_E$ respectively.

\subsubsection{\textbf{\acrfull{iks}}}
It is an online variant of the Kolmogorow-Smirnov (KS) test \cite{Reis2016FastUO}.
The principle behind \acrshort{iks} detector is to apply the KS test on each individual feature.
The detection of a change in a single feature may be sufficient to trigger the presence of a concept drift.
The window size used is $5 w$ and the confidence level $\alpha$ is $0.01$.
First, we compute the pairwise distances $\bm{d}_{i,j}$ with \eqref{eq:pdist}.
Then, we create the feature vectors $\{\bm{m}^{ref}\}$ and $\bm{m}$, where we use the \textit{mean}, \textit{standard deviation}, \textit{maximum} and \textit{minimum} as statistical features, respectively on the reference and test-set.
We use those feature vectors derived from the embedding representation with the \acrshort{iks} detection method.

\subsection{Datasets and preprocessing}
\label{sec:dataprep}

We conduct our experiments on two synthetic and seven real-world datasets. 
A summary of the characteristics of the used datasets is reported in Table~\ref{tab:data_info}.

The synthetic datasets \textit{RBF} and \textit{MovingRBF} are created using the \verb|make_classification| method of the 
scikit-learn Python toolbox \cite{scikit-learn}. For both datasets, we create four clusters with unit variance in a ten dimensional feature space.
For the  \textit{RBF} dataset we add ten more features to each feature vector, where  five are redundant (i.e.\ copies of the existing informative features) and five are just white noise. 

Additionally, we consider seven real-world datasets which are already utilized in others drift detection works \cite{Sethi2017OnTR,Sethi2016MonitoringCB}.
They are binary classification problems and the data instances were shuffled randomly, in order to remove any bias or drifts present in the data. 
All the datasets were processed to have only numeric and binary values, and normalized to have each attribute zero mean and unit standard deviation.

The initial $50\%$ of each dataset is assumed to be labeled, and is used to train the model.
The last half of the data is considered as an unsupervised data stream, and it is used for testing. 
The first $25\%$ of the test data is used as reference dataset to initialize the statistics for the drift detectors. 
A concept drift is induced in the last $50\%$ of the test-stream.

\begin{table}[tb]
    \centering
    \caption{Characteristics of datasets used.}
    \begin{tabular*}{\linewidth}{l @{\extracolsep{\fill}} l l l}
    \toprule
    \textbf{Dataset} & \textbf{\#Instances} & \textbf{\#Attributes} & \textbf{\#Classes} \\
    \midrule
        \textit{adult} & 48842 & 65 & 2\\
        \textit{bank} & 45211 & 48 & 2\\
        \textit{digits08} & 1499 & 16 & 2\\
        \textit{digits17} & 1557 & 16 & 2\\
        \textit{musk} & 6598 & 166 & 2\\
        \textit{phishing} & 11055 & 46 & 2\\
        \textit{wine} & 6497 & 12 & 2\\
        \midrule
        \textit{RBF} & 10000 & 20 & 4\\
        \textit{MovingRBF} & 10000 & 10 & 4\\
    \bottomrule
    \end{tabular*}
    \label{tab:data_info}
\end{table}

\subsection{Inducing concept drift}

We artificially introduce concept drift after $50 \%$ of the test data stream.
By controlling the location and nature of the drift, it is possible to evaluate the drift detection capabilities of different methods.
The drift induction process is a standard benchmark framework for drift detection \cite{Sethi2016MonitoringCB,Sethi2017OnTR}.
The features are corrupted based on their importance to the classification task to create a task-dependent drift.
First, the features are ranked based on their information gain \cite{Sethi2016MonitoringCB}, and then either the most or least informative features are selected, by choosing the top $25 \%$ (\textit{most informative}) or the bottom $25 \%$ (\textit{least informative}) of features from the ranked list.

We focus on both \textit{step} and \textit{gradual} drifts.
The \textit{step} drift is induced by randomly shuffling the values of a subset of features \cite{Sethi2016MonitoringCB}.
This approach ensures that feature drifts are induced while also maintaining the original data properties of the dataset.
The \textit{gradual} drift is induced by corrupting a subset of features with noise $\eta \sim \mathcal{N}(\mu,\,\sigma^2)$.
While $\mu = 1$, the value of $\sigma$ increases over time, from $\sigma=0$ at $50\%$ to $\sigma=2$ after $75 \%$ of the data stream.
This corruption strategy simulates a gradual mean shift of the features.
Differently from \cite{Sethi2016MonitoringCB}, we corrupt the samples of all classes, instead of picking only one class.
In this way, the drift has bigger impact on the classification task, and also could be relevant when modifying the least important features. 

Only in the \textit{MovingRBF} dataset, we induce the drift by randomly re-sampling the location of the class centroids, and then moving them from the initial to this final position when generating the data, either in a step or gradual manner.
In this way, we directly affect the position of the class centroids, while in the other drift induction strategies, it is a consequence of the corruption of the attributes.

\subsection{Implementation details}
We use the same architecture and hyper-parameters for all experiments.
The encoder is a multilayer perceptron (MLP) neural network. 
It consists of three hidden layers of 256, 64 and 3 neurons.
Hence, the embedding represntation has dimension $d=3$  ($f_e(\bm{x}) \in \mathbb{R}^3$).
The classifier is a single fully connected layer with the number of output neurons equal to the number of classes.
The ReLU non-linear function is used for the hidden layers and the softmax activation for the output.
The optimizer used is SGD with 0.9 of momentum. 
Also, a dropout of 0.25 and $L_2$ norm penalization with weight 0.001 is employed in order to reduce generalization error.

Unless explicitly reported, the default value of the drift detection history, maximum delay and detection ratio are $w=50$, $d_{max}=6 w$, and $r=0.25$ respectively.

The source code used in the experiments is publicly available on \href{http://github.com/Castel44/TSDD}{github.com/Castel44/TSDD}

\subsection{Evaluation measures}

\begin{figure}[tb]
    \centering
    \includegraphics[width=0.95\linewidth]{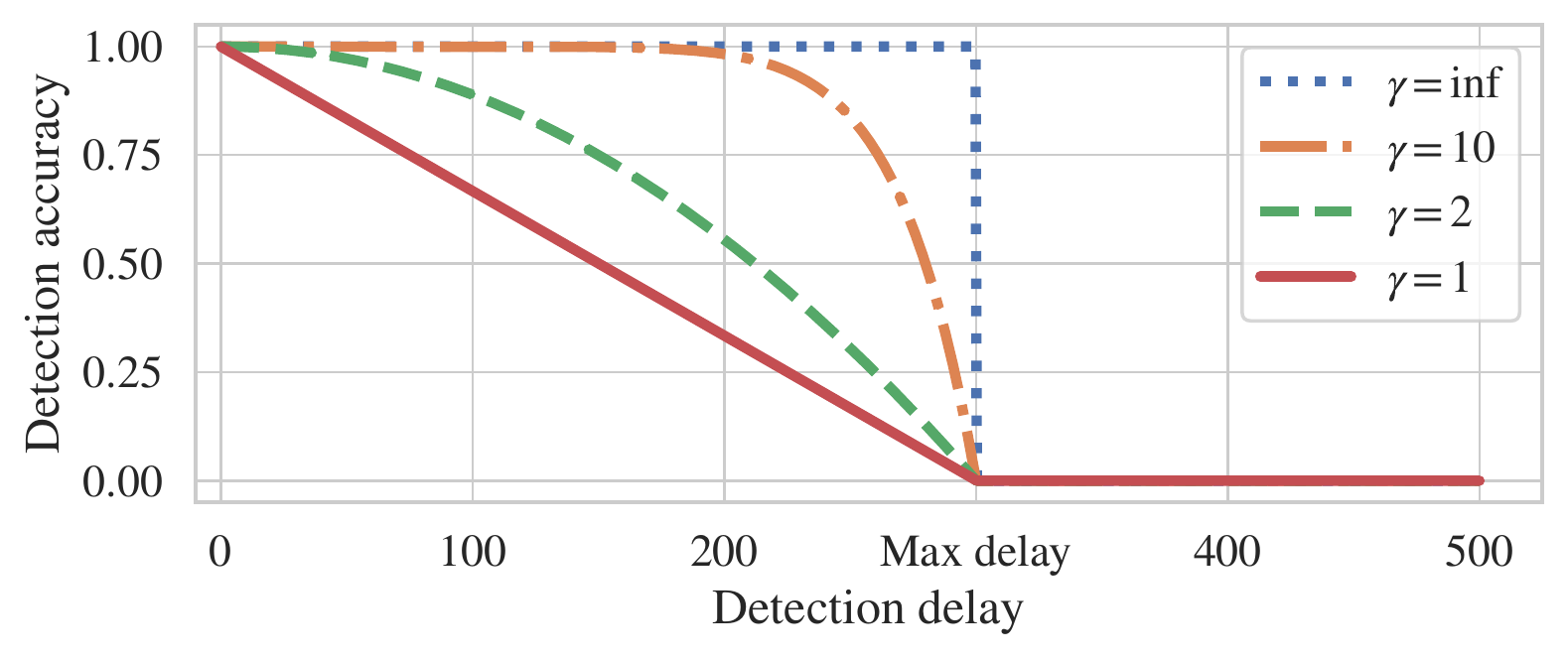}
    \caption{Penalized drift detection accuracy $\widehat{DA}$.}
    \label{fig:acc_mod}
\end{figure}

A dataset exhibits \textit{real drift}, if the performance of the classifier is significantly degraded by the drift.
With the generalization error defined as $GE= | Acc_{Train} - Acc_{Valid} |$, we attribute a real drift to a dataset  if the classification accuracy drops below the validation accuracy with the generalization error as margin, i.e.\ if
\begin{equation}\label{eq:GE}
    Acc_{Test} < Acc_{Valid} - GE \,.
\end{equation}
If the performance after the drift does not decrease below this threshold, a \textit{virtual drift} occurs which should not be detected.

The performance of  drift detectors are assessed by Detection Accuracy $DA$, the False Alarm Rate (False Positive Rate $FPR$) and the Detection Delay $d$ \cite{Sethi2016MonitoringCB,Lu2019LearningUC}.
The $DA$ is a binary value that signals the correct detection of the drift event.
Note that, in case of virtual drift, the goal is to not have any detection, i.e. any detection is considered a false positive.
In case of real drift, we account for the detection delay 
by penalizing $DA$ with a term proportional to the  delay $d$,
\begin{equation}
    \widehat{DA} = 
    \begin{cases}
    DA - \Big( \frac{d}{d_{max}} \Big)^\gamma & \text{for real drift} \\
    DA & \text{for virtual drift}
    \end{cases}\,,
\end{equation}
where $d_{max}$ is the maximum acceptable delay which is set to $d_{max}=300$ samples and $\gamma>0$ a free parameter. Thus, an earlier detection of real drift is better than a later detection. 
Fig.~\ref{fig:acc_mod} shows the penalized drift accuracy $\widehat{DA}$ for various values of $\gamma$. The default value is $\gamma=2$.

In order to characterize the performance with a single  metric, we propose an accumulation of those performance values.
With the true negative rate \textit{TNR} = 1-\textit{FPR}, we define the metric $H$ as the harmonic mean between $\widehat{DA}$ and \textit{TNR},
\begin{equation}
    H = 2 \cdot \frac{\widehat{DA} \cdot TNR}{\widehat{DA} + TNR}
\end{equation}
The $H$ metric is continuous and bounded $H \in [0, 1]$. $H=1$ indicates perfect detection and no false alarms, and $H=0$ means that either the drift has not been detected or the false alarm rate is 1.

All  experiments have been repeated 10 times with different random seeds.
In order to assess the statistical significance of the results, we use the Friedman non-parameteric test with 0.05 confidence level, followed by Nemenyi post-hoc test, see \cite{demsarCD06}.
This evaluation approach allows the simultaneous comparison of different methods considering several data sets.
To visualize the results, we use the critical difference diagram \cite{demsarCD06}, where a thick horizontal line shows the algorithms that are not significantly different in terms of $H$-score.

\section{Results and discussion}

In our experiments, the goal is to pinpoint the single drift event, in case of \textit{real concept drift}, and to not detect any change in case of \textit{virtual drift}, where the changes in the data does not have an impact on the predictions.
The effect of corrupting the most and least informative feature is reported in Table~\ref{tab:accuracies}.
The similarity of train and validation accuracy indicates an initial static dataset.
We report real drift in all the datasets when the most informative features are corrupted.
Instead, the corruption of the least informative features induces real drift only on \textit{digits08} (only step drift) and \textit{phishing} datasets. While, the others, reports virtual drift.

We empirically validate the effect of the constraint embedding module on the latent representation by using the generalized variance (GV) \cite{gvwilks1932} of the validation data around its class centroids.
The GV is proportional to the area of the ellipsoid in the $d$-dimensional space \cite{anderson1962introduction}, and it includes meaningful information about the data sparsity.
The effect of the constraining embedding module is to reduce the GV of the latent representation by $63.31\%$ to $99.95\%$  depending on the dataset. 
Therefore, the constraint module leads to a strong concentration of data samples around the centroids as desired.

We compare the performance of the proposed drift detection framework with the algorithms \acrshort{ema} and \acrshort{zscore} against different SotA detection algorithms.
The performance is evaluated by using the proposed constrained embedding module  (indicated by \cmark), which returns the class centroids at the end of the training, and compared to the case where the constraints module was not used during training (indicated by \xmark), where the class centroids need to be calculated after the training, as the average of the training data of each class.
Additionally, we also compare to the SotA for unsupervised concept drift detection, with an analysis directly on the input data distribution with the HDDDM$_I$ algorithm.

We report the drift detection results  for a  representative subset of the all the studied datasets in Table~\ref{tab:results}. Result of all datasets can be found in the supplementary material \cite{castellaniConcept2021}.
By analyzing all the results, we observe that the proposed framework with the inclusion of the constraint embedding module outperforms the SotA counterpart in terms of $H$-score in 79 experiments, has the same performance level in 30 experiments, and is worse in 27 out of 136 experiments. 
Moreover, the proposed framework always shows comparable or better performance than the unsupervised drift detection on the raw input data, HDDDM$_I$, which is not able to deal with virtual drifts as expected.
All methods successfully detect the drift on the synthetic dataset \textit{MovRBF}. Notice that there is no distinction between least and most important features in that dataset, as described in Section \ref{sec:dataprep}.

In the Table~\ref{tab:results}, we can  observe the advantage of the cumulative measure $H$, which accounts for all three performance measures \textit{DA}, \textit{TNR} and \textit{Delay}.
For example, comparing the performance of  the \textit{IKS} algorithm on the \textit{wine} dataset with step drift on the most important features, the performance is almost perfect for \textit{DA} and \textit{TNR}, but the delay is rather large when trained without the constraints on the embedding  (indicated with \xmark). 
This is reflected in a reduced performance measure $H$. In comparison, including   the constraint (\cmark) also gives almost perfect performance on \textit{DA} and \textit{TNR}, but reduces the delay considerably, which is reflected in an improved $H$ measure.   
In that way, the $H$ measure gives a more comprehensive picture of the performance as relying on one single metrics.
Also, from that example, it is clear that the inclusion of the constrained embedding module improves the performance.

\begin{figure}[t]
    \centering
    \includegraphics[width=1\linewidth]{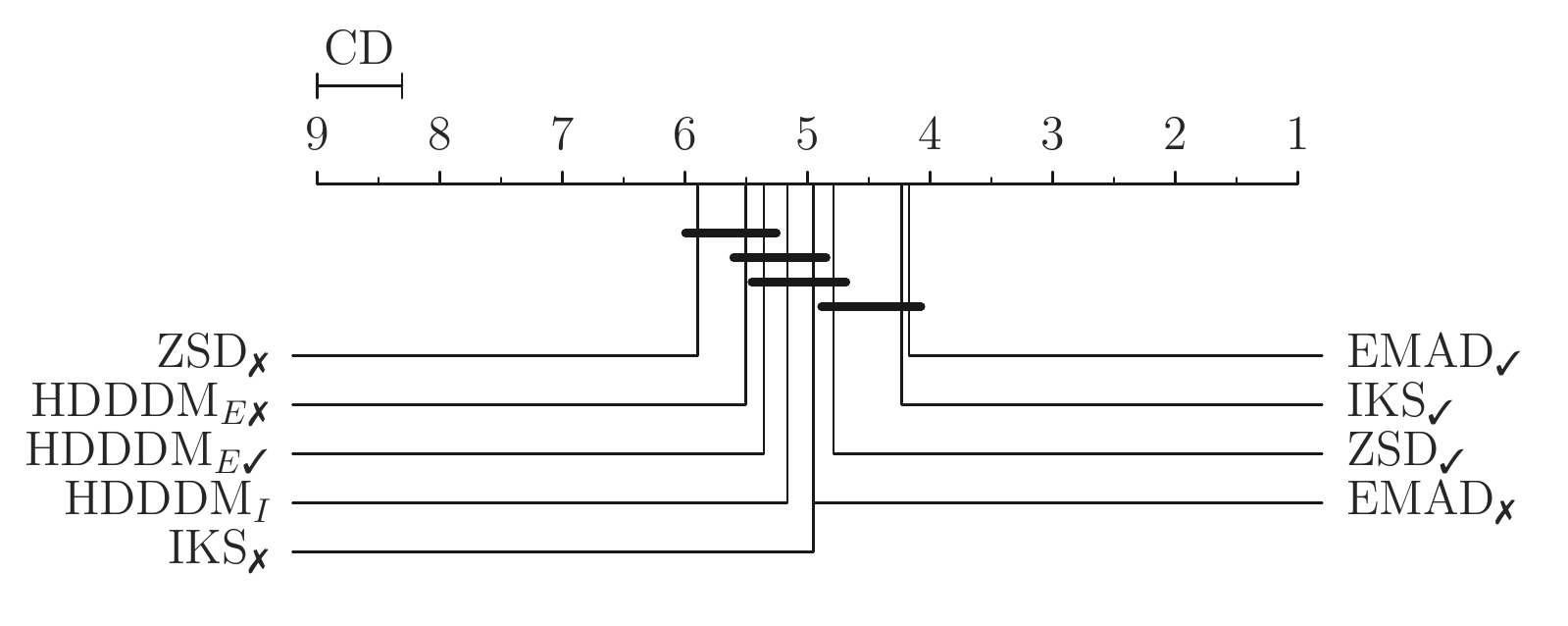}
    \caption{Critical difference diagram based on the $H$-score accumulated over all datasets and drift types showing pairwise statistical difference comparison of different drift detectors. 
    }
    \label{fig:cd_all}
\end{figure}

In Fig.~\ref{fig:cd_all} we report the critical differences based on the $H$-score of the investigated algorithms accumulated over all the datasets and drift types.
The algorithms using the proposed constrained embedding framework (subscript~\cmark) are in all cases, except for HDDDM$_E$, significantly better than the corresponding method without the constraint embedding (subscript~\xmark).
However, there is no single detector that outperform all the others since between the proposed \acrshort{ema}$_{\text{\cmark}}$, \acrshort{zscore}$_{\text{\cmark}}$ and \acrshort{iks}$_{\text{\cmark}}$ the difference is not statistically significant.

\begin{figure}[t]
    \centering
    \includegraphics[width=1\linewidth]{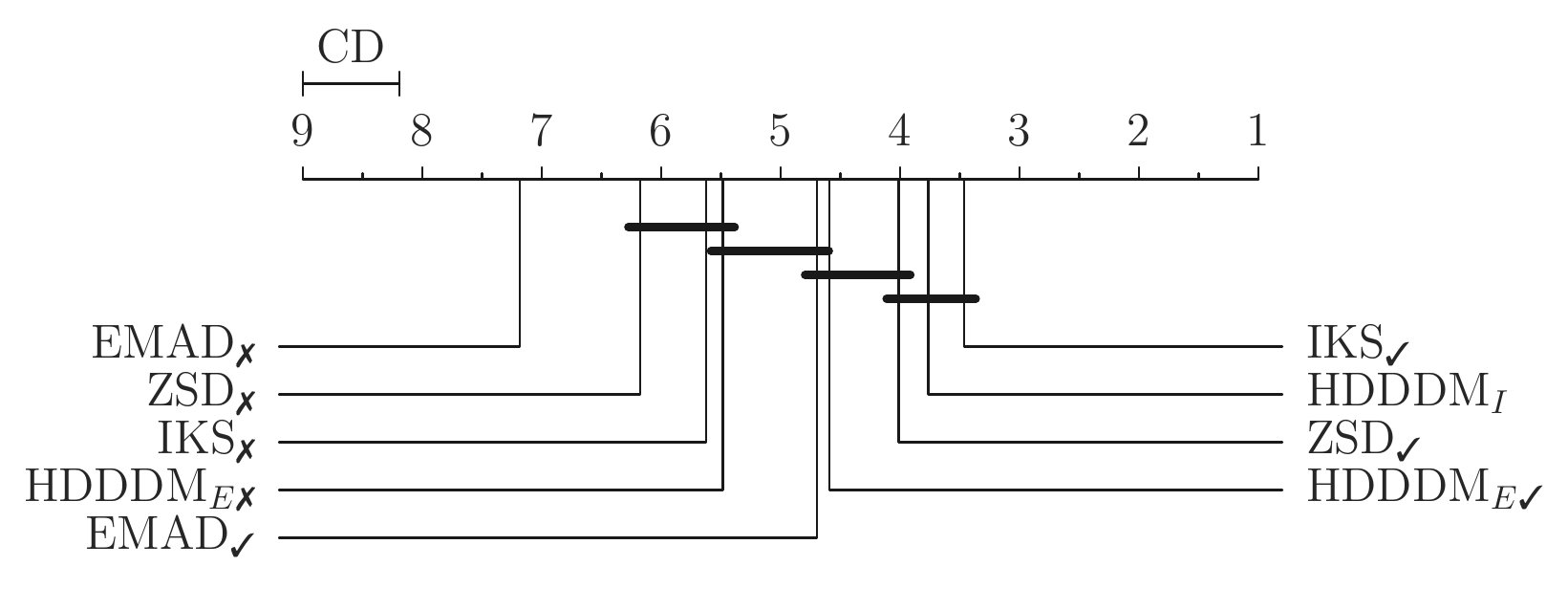}
    \caption{Critical difference diagram based on the $H$-score accumulated over the datasets showing \textit{real drift} only.
    }
    \label{fig:cd_real}
\end{figure}

\begin{figure}[t]
    \centering
    \includegraphics[width=1\linewidth]{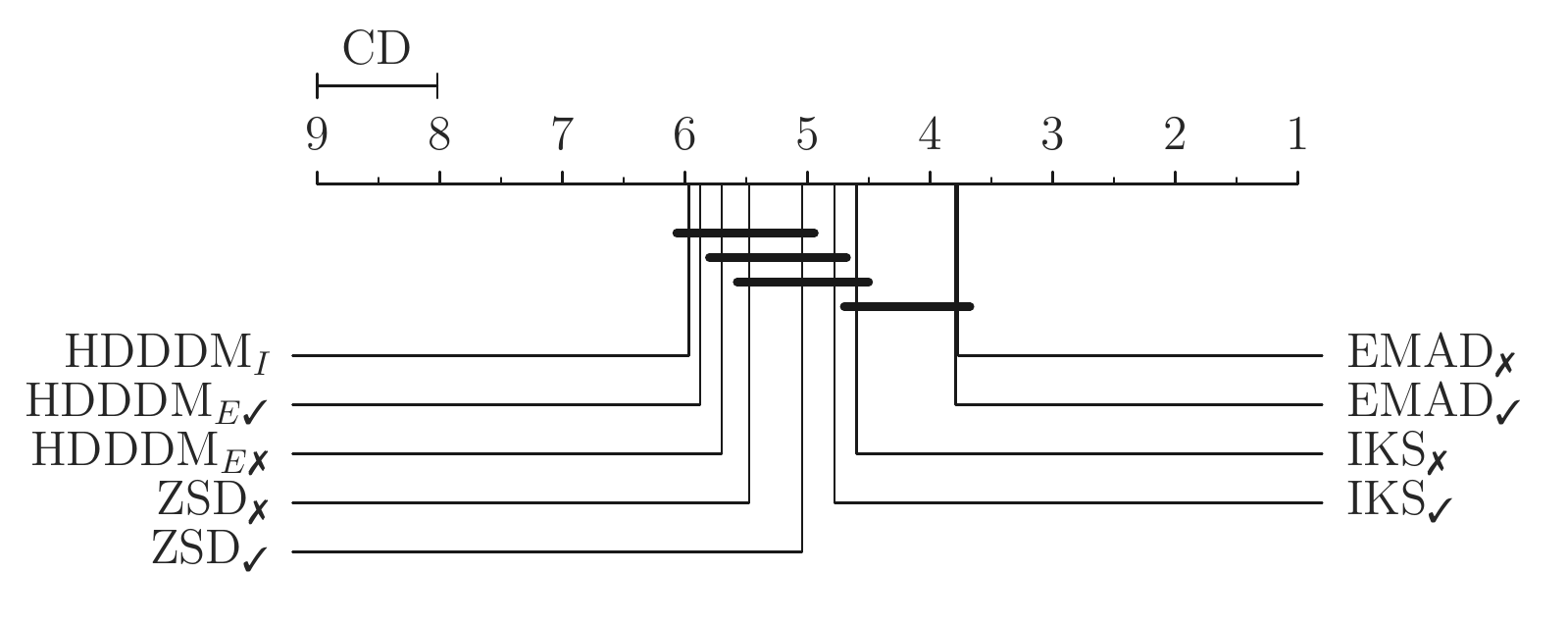}
    \caption{Critical difference diagram based on the $H$-score accumulated over the datasets showing \textit{virtual drift} only.
    }
    \label{fig:cd_virtual}
\end{figure}

For the experiments on datasets showing real drift as reported in Fig.~\ref{fig:cd_real} the trend is even clearer.  
All approaches using the constraints on the embedding are significantly better than the corresponding approaches without the constraints.  
Only HDDDM$_I$ working on the raw input shows a similar performance as the proposed constrained embedding approaches.
However, the performance of HDDDM$_I$ is the worst on the datasets showing virtual drift as it can be expected (see Fig.~\ref{fig:cd_virtual}). 
For the other approaches working on the low dimensional embedding, there is no difference between methods with and without constrains, because virtual drift does not affect the model performance and the latent representation substantially. 

\begin{table}[tb]
\setlength{\tabcolsep}{2.5pt}

    \centering
    \scriptsize
    \caption{Classification accuracies of train, validation, and test-set when the most and least informative features are corrupted.
    Cases  of virtual drift are highlighted.}
    \begin{tabular*}{\linewidth}{l  @{\extracolsep{\fill}} c c |c c |c c}
        \toprule
        \multirow{2}{*}{\textbf{Dataset}} & \multirow{2}{*}{\textbf{Train}} & \multirow{2}{*}{\textbf{Valid}} & \multicolumn{2}{c|}{\textbf{\textit{Step drift}}} & \multicolumn{2}{c}{\textbf{\textit{Gradual drift}}}\\
        & & & \textit{Most} & \textit{Least} & \textit{Most} & \textit{Least} \\
        \midrule
\textit{adult} & 88.8\rpm0.1 & 86.0\rpm0.2 & 77.4\rpm3.4 & \textbf{84.9\rpm0.1} & 72.3\rpm0.8 & \textbf{84.8\rpm0.1} \\
\textit{bank} & 95.3\rpm0.1 & 91.5\rpm0.2 & 77.4\rpm2.8 & \textbf{90.7\rpm0.1} & 81.2\rpm0.5 & \textbf{89.5\rpm0.3} \\
\textit{digits08} & 100.0\rpm0.0 & 99.6\rpm0.3 & 96.3\rpm3.2 & 96.7\rpm1.2  & 98.5\rpm0.6 & \textbf{99.4\rpm0.1} \\
\textit{digits17} & 100.0\rpm0.0 & 100.0\rpm0.0 & 97.5\rpm0.5 & \textbf{99.1\rpm0.4} & 90.9\rpm0.7 & \textbf{99.4\rpm0.2} \\
\textit{musk} & 100.0\rpm0.0 & 98.8\rpm0.1 & 93.6\rpm1.6 & \textbf{98.1\rpm0.4} & 80.6\rpm0.2 & \textbf{89.4\rpm1.0} \\
\textit{phishing} & 98.8\rpm0.0 & 95.0\rpm0.4 & 76.7\rpm4.6 & 90.4\rpm0.5 & 88.9\rpm0.4 & 90.8\rpm0.3 \\
\textit{wine} & 100.0\rpm0.0 & 100.0\rpm0.0 & 80.8\rpm5.7 & \textbf{100.0\rpm0.0}  & 89.7\rpm0.4 & \textbf{100.0\rpm0.0}\\
\midrule
\textit{RBF} & 99.9\rpm0.1 & 99.1\rpm0.5 & 93.5\rpm3.4 & \textbf{99.1\rpm0.6} & 90.8\rpm4.1 & \textbf{98.9\rpm0.6} \\
\textit{MovRBF} & 100.0\rpm0.0 & 100.0\rpm0.0 & 49.9\rpm9.0 & - & 52.4\rpm7.7 & - \\
        \bottomrule
    \end{tabular*}
    \label{tab:accuracies}
\end{table}


\begin{sidewaystable*}[p]
    \scriptsize
    \centering
    \setlength{\tabcolsep}{1.6pt}
    \renewcommand{\arraystretch}{0.8}
    \aboverulesep = 0.605mm 
    \belowrulesep = 0.984mm 
    \caption{Drift detection results. The presence (or not) of the constraint module is reported by $\cmark$ (or $\xmark$).
    Highlighted is the best performing in the comparison with and without constraint module.
    }
    \label{tab:results}
    \begin{tabular*}{\linewidth}{l @{\extracolsep{\fill}} c c c c c c !{\vrule width 0.95pt} c c c c !{\vrule width 0.95pt} c c c c !{\vrule width 0.95pt} c c c c}
\toprule
\multirow{2}{*}{\textbf{Dataset}} & \multirow{2}{*}{\textbf{Detector}} & \multirow{2}{*}{$\xmark$/$\cmark$} & \multicolumn{4}{c!{\vrule width 0.95pt}}{\textbf{\textit{Step drift: most important}}} & \multicolumn{4}{c!{\vrule width 0.95pt}}{\textbf{\textit{Step drift: least important}}} & \multicolumn{4}{c!{\vrule width 0.95pt}}{\textbf{\textit{Gradual drift: most important}}} & \multicolumn{4}{c}{\textbf{\textit{Gradual drift: least important}}}\\
& & & \textit{DA} & \textit{TNR} & \textit{Delay} & \textit{H} & \textit{DA} & \textit{TNR} & \textit{Delay} & \textit{H} & \textit{DA} & \textit{TNR} & \textit{Delay} & \textit{H}  & \textit{DA} & \textit{TNR} & \textit{Delay} & \textit{H} \\

\midrule
\multirow{12}{*}{\textit{digits08}} & \multirow{2}{*}{ZSD} & \xmark & \textbf{1.00\rpm0.00} &  0.94\rpm0.14 & 11.80\rpm16.34 & \textbf{0.96\rpm0.09} & 0.50\rpm0.53 &  0.99\rpm0.04 & - & 0.49\rpm0.51 & 0.60\rpm0.52 &  0.99\rpm0.04 & 14.20\rpm25.53 & 0.60\rpm0.52  & 0.30\rpm0.48 &  1.00\rpm0.00 & 78.70\rpm47.96 & 0.30\rpm0.48  \\ 
& & \cmark & 0.90\rpm0.32 &  \textbf{1.00\rpm0.00} & 7.10\rpm3.96 & 0.90\rpm0.32  & \textbf{1.00\rpm0.00} &  1.00\rpm0.00 & - & \textbf{1.00\rpm0.00} & 0.60\rpm0.52 &  1.00\rpm0.00 & \textbf{9.10\rpm2.38} & 0.60\rpm0.52 & \textbf{0.60\rpm0.52} &  1.00\rpm0.00 & \textbf{19.00\rpm46.24} & \textbf{0.60\rpm0.52} \\ 
\cmidrule{2-19}
& \multirow{2}{*}{EMAD} & \xmark & 0.90\rpm0.32 &  1.00\rpm0.00 & 23.00\rpm38.27 & 0.90\rpm0.32  & 0.50\rpm0.53 &  1.00\rpm0.00 & - & 0.50\rpm0.52 & 0.50\rpm0.53 &  1.00\rpm0.00 & 12.20\rpm29.35 & 0.50\rpm0.53  & 0.60\rpm0.52 &  1.00\rpm0.00 & \textbf{0.00\rpm0.00} & \textbf{0.60\rpm0.52}  \\ 
& & \cmark & 0.90\rpm0.32 &  1.00\rpm0.00 & \textbf{9.50\rpm3.54} & 0.90\rpm0.32  & \textbf{1.00\rpm0.00} &  1.00\rpm0.00 & - & \textbf{1.00\rpm0.01} & 0.60\rpm0.52 &  1.00\rpm0.00 & 12.00\rpm2.62 & 0.60\rpm0.52  & 0.40\rpm0.52 &  1.00\rpm0.00 & 99.30\rpm47.51 & 0.39\rpm0.50 \\ 
\cmidrule{2-19}
& \multirow{2}{*}{IKS} & \xmark & 0.60\rpm0.52 &  1.00\rpm0.00 & 92.10\rpm60.28 & 0.60\rpm0.51  & 0.60\rpm0.52 &  1.00\rpm0.00 & - & 0.57\rpm0.49 & 0.60\rpm0.52 &  1.00\rpm0.00 & \textbf{76.50\rpm7.92} & 0.60\rpm0.51  & 0.30\rpm0.48 &  1.00\rpm0.00 & 105.30\rpm20.68 & 0.30\rpm0.48 \\ 
& & \cmark & \textbf{0.90\rpm0.32} &  1.00\rpm0.00 & \textbf{41.20\rpm9.89} & \textbf{0.90\rpm0.32}  & \textbf{1.00\rpm0.00} &  1.00\rpm0.00 & - & \textbf{0.99\rpm0.00} & 0.70\rpm0.48 &  1.00\rpm0.00 & 108.30\rpm96.02 & 0.64\rpm0.45 & \textbf{0.70\rpm0.48} &  1.00\rpm0.00 & \textbf{20.70\rpm65.46} & \textbf{0.68\rpm0.47} \\ 
\cmidrule{2-19}
& \multirow{2}{*}{HDDDM$_{E}$} & \xmark & 0.50\rpm0.53 &  1.00\rpm0.00 & 145.00\rpm0.00 & 0.47\rpm0.50  & 0.70\rpm0.48 &  1.00\rpm0.00 & - & 0.63\rpm0.44 & 0.40\rpm0.52 &  1.00\rpm0.00 & 131.00\rpm46.06 & 0.38\rpm0.48  & \textbf{0.50\rpm0.53} &  1.00\rpm0.00 & 145.30\rpm59.23 & \textbf{0.48\rpm0.50} \\ 
& & \cmark & \textbf{0.80\rpm0.42} &  1.00\rpm0.00 & 130.60\rpm45.89 & \textbf{0.75\rpm0.40}  & \textbf{1.00\rpm0.00} &  1.00\rpm0.00 & - & \textbf{0.94\rpm0.00} & 0.50\rpm0.53 &  1.00\rpm0.00 & 145.90\rpm59.64 & 0.45\rpm0.48  & 0.30\rpm0.48 &  1.00\rpm0.00 & 145.00\rpm0.00 & 0.28\rpm0.45 \\ 
\cmidrule{2-19}
&HDDDM$_{I}$& -  & 0.50\rpm0.53 &  1.00\rpm0.53 & 158.00\rpm21.28 & 0.45\rpm0.48  & 0.40\rpm0.52 &  1.00\rpm0.00 & - & 0.40\rpm0.52 & 0.50\rpm0.53 &  1.00\rpm0.53 & 0.00\rpm0.00 & 0.50\rpm0.53  & 0.60\rpm0.52 &  1.00\rpm0.00 & 0.00\rpm0.00 &  0.60\rpm0.52 \\ 

\specialrule{0.95pt}{0.605mm}{0.984mm} 
\multirow{12}{*}{\textit{phishing}} & \multirow{2}{*}{ZSD} & \xmark & \textbf{1.00\rpm0.00} &  0.08\rpm0.04 & \textbf{0.00\rpm0.00} & 0.14\rpm0.07  & 0.50\rpm0.53 &  0.44\rpm0.29 & 261.10\rpm331.56 & 0.30\rpm0.38 & 0.70\rpm0.48 &  \textbf{0.39\rpm0.17} & 201.80\rpm203.94 & 0.28\rpm0.26  & 0.40\rpm0.52 &  \textbf{0.58\rpm0.20} & 273.30\rpm377.38 & 0.33\rpm0.43 \\ 
& & \cmark & 0.90\rpm0.32 &  \textbf{0.77\rpm0.21} & 80.80\rpm155.72 & \textbf{0.82\rpm0.28}  & 0.70\rpm0.48 &  0.13\rpm0.07 & 0.00\rpm0.00 & 0.19\rpm0.16 & \textbf{1.00\rpm0.00} &  0.13\rpm0.07 & \textbf{0.00\rpm0.00} & 0.22\rpm0.11  & \textbf{0.90\rpm0.32} &  0.13\rpm0.06 & \textbf{0.00\rpm0.00} & \textbf{0.22\rpm0.12} \\ 
\cmidrule{2-19}
& \multirow{2}{*}{EMAD} & \xmark & 0.10\rpm0.32 &  1.00\rpm0.00 & \textbf{6.30\rpm19.92} & 0.10\rpm0.31  & 0.30\rpm0.48 &  1.00\rpm0.00 & 0.00\rpm0.00 & 0.30\rpm0.48 & 0.00\rpm0.00 &  1.00\rpm0.00 & 0.00\rpm0.00 & 0.00\rpm0.00  & 0.10\rpm0.32 &  1.00\rpm0.00 & 0.00\rpm0.00 & 0.10\rpm0.32 \\ 
& & \cmark & \textbf{0.70\rpm0.48} &  1.00\rpm0.00 & 197.50\rpm214.28 & \textbf{0.66\rpm0.46}  & \textbf{0.70\rpm0.48} &  1.00\rpm0.00 & 0.00\rpm0.00 & \textbf{0.70\rpm0.48} & \textbf{0.20\rpm0.42} &  1.00\rpm0.00 & 0.00\rpm0.00 & \textbf{0.20\rpm0.42}  & \textbf{0.90\rpm0.32} &  1.00\rpm0.00 & 0.00\rpm0.00 & \textbf{0.90\rpm0.32} \\ 
\cmidrule{2-19}
& \multirow{2}{*}{IKS} & \xmark & 1.00\rpm0.00 &  0.71\rpm0.09 & 9.20\rpm15.96 & 0.83\rpm0.06  & 0.30\rpm0.48 &  0.66\rpm0.05 & \textbf{0.00\rpm0.00} & 0.24\rpm0.38 & 0.70\rpm0.48 &  0.68\rpm0.09 & 135.25\rpm56.36  & 0.55\rpm0.38  & 0.20\rpm0.42 &  0.70\rpm0.07 & \textbf{0.00\rpm0.00} & 0.17\rpm0.36 \\ 
& & \cmark & 1.00\rpm0.00 &  0.78\rpm0.12 & 16.60\rpm35.15 & 0.87\rpm0.08  & \textbf{0.60\rpm0.52} &  0.73\rpm0.14 & 44.80\rpm39.99 & \textbf{0.51\rpm0.44} & 0.70\rpm0.48 &  0.73\rpm0.17 & 149.40\rpm53.25 & 0.50\rpm0.36  & \textbf{0.70\rpm0.48} &  0.76\rpm0.08 & 186.20\rpm438.38 & \textbf{0.59\rpm0.41}  \\ 
\cmidrule{2-19}
& \multirow{2}{*}{HDDDM$_{E}$} & \xmark & 1.00\rpm0.00 &  0.82\rpm0.04 & 111.10\rpm39.41 & 0.87\rpm0.05  & 0.70\rpm0.48 &  0.81\rpm0.03 & 403.30\rpm606.53 & 0.63\rpm0.43 & 0.30\rpm0.48 &  0.78\rpm0.04 & 558.20\rpm402.23 & 0.06\rpm0.17  & \textbf{0.90\rpm0.32} &  0.81\rpm0.06 & 164.80\rpm408.80 & \textbf{0.80\rpm0.28} \\ 
& & \cmark & 1.00\rpm0.00 &  0.78\rpm0.03 & 132.40\rpm42.85 & 0.83\rpm0.06  & 0.40\rpm0.52 &  0.77\rpm0.03 & 146.20\rpm386.87 & 0.32\rpm0.42 & 0.20\rpm0.42 &  0.79\rpm0.03 & 443.30\rpm349.13 & 0.16\rpm0.34  & 0.20\rpm0.42 &  0.75\rpm0.04 & 369.90\rpm593.49 & 0.17\rpm0.35 \\ 
\cmidrule{2-19}
&HDDDM$_{I}$& -  & 1.00\rpm0.00 &  0.84\rpm0.00 & 14.30\rpm2.50 & 0.91\rpm0.00  & 0.30\rpm0.48 &  0.84\rpm0.00 & 17.70\rpm4.00 & 0.27\rpm0.44 & 0.80\rpm0.42 &  0.84\rpm0.42 & 12.00\rpm0.00 & 0.73\rpm0.38 & 0.10\rpm0.32 &  0.84\rpm0.00 & 11.00\rpm0.00 & 0.09\rpm0.29  \\ 

\specialrule{0.95pt}{0.605mm}{0.984mm} 
\multirow{12}{*}{\textit{wine}} & \multirow{2}{*}{ZSD} & \xmark & 1.00\rpm0.00 &  0.67\rpm0.13 & 19.30\rpm21.97 & 0.80\rpm0.10  & 0.60\rpm0.52 &  0.72\rpm0.19 & - & 0.55\rpm0.47 & 1.00\rpm0.00 &  0.79\rpm0.10 & 18.50\rpm12.26 & 0.88\rpm0.06  & 0.10\rpm0.32 &  0.63\rpm0.18 & - & 0.09\rpm0.30 \\ 
& & \cmark & 1.00\rpm0.00 &  \textbf{1.00\rpm0.00} & 28.30\rpm13.15 & \textbf{1.00\rpm0.01}  & \textbf{1.00\rpm0.00} &  \textbf{1.00\rpm0.00} & - & \textbf{1.00\rpm0.00} & 1.00\rpm0.00 &  \textbf{1.00\rpm0.00} & 99.80\rpm11.43 & \textbf{0.98\rpm0.01}  & \textbf{1.00\rpm0.00} &  \textbf{1.00\rpm0.00} & - & \textbf{1.00\rpm0.00}\\ 
\cmidrule{2-19}
& \multirow{2}{*}{EMAD} & \xmark & 0.00\rpm0.00 &  \textbf{1.00\rpm0.00} & \textbf{0.00\rpm0.00} & 0.00\rpm0.00  & 1.00\rpm0.00 &  1.00\rpm0.00 & - & 1.00\rpm0.00 & 1.00\rpm0.00 &  1.00\rpm0.00 & 68.30\rpm64.54 & 0.96\rpm0.12  & 1.00\rpm0.00 &  1.00\rpm0.00 & - & 1.00\rpm0.00  \\ 
& & \cmark & \textbf{1.00\rpm0.00} &  0.95\rpm0.02 & 13.50\rpm5.64 & \textbf{0.97\rpm0.01}  & 1.00\rpm0.00 &  0.95\rpm0.02 & - & 0.99\rpm0.01 & 1.00\rpm0.00 &  0.95\rpm0.02 & 41.30\rpm28.06 & 0.97\rpm0.02  & 1.00\rpm0.00 &  0.95\rpm0.03 & - & 0.97\rpm0.03  \\ 
\cmidrule{2-19}
& \multirow{2}{*}{IKS} & \xmark & 0.90\rpm0.18 &  1.00\rpm0.00 & 224.00\rpm54.86 & 0.68\rpm0.21  & 0.30\rpm0.48 &  1.00\rpm0.00 & - & 0.30\rpm0.48 & 1.00\rpm0.00 &  1.00\rpm0.00 & 58.40\rpm8.82 & 1.00\rpm0.00 & 0.30\rpm0.48 &  1.00\rpm0.00 & - & 0.30\rpm0.48\\ 
& & \cmark & 1.00\rpm0.00 &  1.00\rpm0.00 & \textbf{94.20\rpm18.60} & \textbf{0.98\rpm0.01}  & \textbf{0.90\rpm0.32} &  1.00\rpm0.00 & - & \textbf{0.90\rpm0.32} & 1.00\rpm0.00 &  1.00\rpm0.00 & 61.70\rpm25.49 & 0.99\rpm0.01 & \textbf{1.00\rpm0.00} &  1.00\rpm0.00 & - & \textbf{1.00\rpm0.00} \\ 
\cmidrule{2-19}
& \multirow{2}{*}{HDDDM$_{E}$} & \xmark & 1.00\rpm0.00 &  0.88\rpm0.05 & 39.60\rpm16.85 & 0.93\rpm0.03  & 0.50\rpm0.53 &  0.88\rpm0.06 & - & 0.47\rpm0.50 & 1.00\rpm0.00 &  0.88\rpm0.04 & 42.60\rpm9.99 & 0.93\rpm0.02  & 0.30\rpm0.48 &  0.92\rpm0.06 & - & 0.29\rpm0.47  \\ 
& & \cmark & 1.00\rpm0.00 &  0.89\rpm0.08 & 23.40\rpm13.75 & 0.94\rpm0.05  & 0.70\rpm0.48 &  0.90\rpm0.06 & - & 0.66\rpm0.46 & 1.00\rpm0.00 &  0.85\rpm0.05 & 37.80\rpm11.35 & 0.92\rpm0.03  & 0.40\rpm0.52 &  0.89\rpm0.06 & - & 0.37\rpm0.48  \\ 
\cmidrule{2-19}
&HDDDM$_{I}$& -  & 1.00\rpm0.00 &  0.88\rpm0.00 & 22.50\rpm2.42 & 0.93\rpm0.00  & 0.00\rpm0.00 &  0.88\rpm0.00 & - & 0.00\rpm0.00 & 1.00\rpm0.00 &  0.88\rpm0.00 & 44.90\rpm1.97 & 0.93\rpm0.00  & 0.00\rpm0.00 &  0.88\rpm0.00 & - & 0.00\rpm0.00  \\

\specialrule{0.95pt}{0.605mm}{0.984mm} 
\multirow{12}{*}{\textit{RBF}} & \multirow{2}{*}{ZSD} & \xmark & 0.90\rpm0.32 &  0.06\rpm0.04 & \textbf{0.00\rpm0.00} & 0.10\rpm0.08  & 0.10\rpm0.32 &  0.08\rpm0.06 & - & 0.00\rpm0.01& 1.00\rpm0.00 &  0.10\rpm0.10 & 3.80\rpm9.81 & 0.17\rpm0.16  & 0.10\rpm0.32 &  0.08\rpm0.08 & - & 0.01\rpm0.03  \\ 
& & \cmark & \textbf{1.00\rpm0.00} &  \textbf{0.51\rpm0.27} & 17.40\rpm38.19 & \textbf{0.64\rpm0.21}  & \textbf{0.40\rpm0.52} &  \textbf{0.56\rpm0.30} & - & \textbf{0.31\rpm0.42} & 1.00\rpm0.00 &  \textbf{0.49\rpm0.25} & 4.20\rpm6.81 & \textbf{0.62\rpm0.24}  & 0.20\rpm0.42 &  \textbf{0.51\rpm0.24} & - & \textbf{0.14\rpm0.29}  \\ 
\cmidrule{2-19}
& \multirow{2}{*}{EMAD} & \xmark & 0.10\rpm0.32 &  1.00\rpm0.00 & \textbf{0.00\rpm0.00} & 0.10\rpm0.32  & 0.90\rpm0.32 &  1.00\rpm0.00 & - & 0.90\rpm0.32 & 0.60\rpm0.52 &  1.00\rpm0.00 & 190.20\rpm163.24 & 0.45\rpm0.48  & 0.90\rpm0.32 &  1.00\rpm0.00 & - & 0.90\rpm0.32 \\ 
& & \cmark & \textbf{0.90\rpm0.32} &  1.00\rpm0.00 & 90.20\rpm87.50 & \textbf{0.89\rpm0.31}  & 0.80\rpm0.42 &  1.00\rpm0.00 & - & 0.80\rpm0.42 & \textbf{1.00\rpm0.00} &  1.00\rpm0.00 & \textbf{47.00\rpm48.84} & \textbf{0.99\rpm0.02}  & 0.90\rpm0.32 &  1.00\rpm0.00 & - & 0.90\rpm0.32 \\ 
\cmidrule{2-19}
& \multirow{2}{*}{IKS} & \xmark & 0.60\rpm0.52 &  1.00\rpm0.00 & 259.90\rpm128.96 & 0.37\rpm0.40  & 0.80\rpm0.42 &  1.00\rpm0.01 & - & 0.80\rpm0.42 & 0.90\rpm0.32 &  1.00\rpm0.00 & 207.70\rpm142.06 & 0.74\rpm0.37  & \textbf{0.90\rpm0.32} &  0.99\rpm0.04 & - & \textbf{0.89\rpm0.31} \\ 
& & \cmark & \textbf{0.90\rpm0.32} &  0.99\rpm0.04 & 169.00\rpm120.60 & \textbf{0.83\rpm0.30}  & 0.80\rpm0.42 &  0.99\rpm0.03 & - & 0.79\rpm0.42 & \textbf{1.00\rpm0.00} &  1.00\rpm0.00 & \textbf{85.40\rpm33.56} & \textbf{0.98\rpm0.02}  & 0.50\rpm0.53 &  1.00\rpm0.01 & - & 0.50\rpm0.53 \\ 
\cmidrule{2-19}
& \multirow{2}{*}{HDDDM$_{E}$} & \xmark & 0.90\rpm0.32 &  0.88\rpm0.07 & 100.40\rpm50.92 & 0.81\rpm0.29  & 0.80\rpm0.42 &  0.88\rpm0.05 & - & 0.68\rpm0.41 & 1.00\rpm0.00 &  0.88\rpm0.09 & 66.10\rpm34.42 & 0.93\rpm0.05  & 0.60\rpm0.52 &  0.89\rpm0.08 & - & 0.57\rpm0.49 \\ 
& & \cmark & \textbf{1.00\rpm0.00} &  0.87\rpm0.08 & 82.20\rpm24.98 & \textbf{0.92\rpm0.05}  & 0.50\rpm0.53 &  0.90\rpm0.10 & - & 0.47\rpm0.49 & 1.00\rpm0.00 &  0.91\rpm0.07 & 50.10\rpm25.91 & 0.95\rpm0.04  & 0.50\rpm0.53 &  0.93\rpm0.09 & - & 0.46\rpm0.49 \\ 
\cmidrule{2-19}
&HDDDM$_{I}$& -  & 0.90\rpm0.32 &  0.89\rpm0.32 & 165.10\rpm122.68 & 0.65\rpm0.37  & 0.70\rpm0.48 &  0.89\rpm0.08 & - & 0.65\rpm0.45 & 1.00\rpm0.00 &  0.89\rpm0.00 & 41.80\rpm30.26 & 0.94\rpm0.04  & 0.10\rpm0.32 &  0.89\rpm0.08 & - & 0.10\rpm0.31 \\ 

\specialrule{0.95pt}{0.605mm}{0.984mm} 
\multirow{12}{*}{\textit{MovRBF}} & \multirow{2}{*}{ZSD} & \xmark & 1.00\rpm0.00 &  0.82\rpm0.18 & 2.50\rpm2.07 & 0.89\rpm0.11 & - & -& - & - & 1.00\rpm0.00 &  0.86\rpm0.00 & \textbf{5.80\rpm4.05} & 0.89\rpm0.22 & - & -& -& -\\ 
& & \cmark & 1.00\rpm0.00 & \textbf{1.00\rpm0.00} & 4.10\rpm0.32 & \textbf{1.00\rpm0.00} & - & -& -& - & 1.00\rpm0.00 &  \textbf{1.00\rpm0.00} & 65.40\rpm40.07 & \textbf{0.99\rpm0.01} & - & -& -& -\\ 
\cmidrule{2-19}
& \multirow{2}{*}{EMAD} & \xmark & 1.00\rpm0.00 &  1.00\rpm0.00 & 6.20\rpm1.75 & 1.00\rpm0.00 & - & -& -& - & 1.00\rpm0.00 &  1.00\rpm0.00 & 23.10\rpm28.78 & \textbf{1.00\rpm0.01} & - & -& -& - \\ 
& & \cmark  & 1.00\rpm0.00 &  1.00\rpm0.00 & 4.00\rpm0.00 &1.00\rpm0.00 & - & -& -& - & 1.00\rpm0.00 &  1.00\rpm0.00 & 17.70\rpm9.87 & 1.00\rpm0.00 & - & -& -& -\\ 
\cmidrule{2-19}
& \multirow{2}{*}{IKS} & \xmark & 1.00\rpm0.00 &  1.00\rpm0.00 & 57.20\rpm34.52 & 0.99\rpm0.01 & - & -& -& - & 1.00\rpm0.00 &  1.00\rpm0.00 & 77.70\rpm30.74 & 0.99\rpm0.02 & - & -& -& -  \\ 
& & \cmark & 1.00\rpm0.00 &  0.99\rpm0.04 & \textbf{29.30\rpm15.17} & 0.99\rpm0.02 & - & -& -& - & 1.00\rpm0.00 &  1.00\rpm0.00 & 50.80\rpm22.46 & 1.00\rpm0.00 & - & -& -& -\\ 
\cmidrule{2-19}
& \multirow{2}{*}{HDDDM$_{E}$} & \xmark & 1.00\rpm0.00 &  1.00\rpm0.00 & 7.40\rpm1.84 & 1.00\rpm0.00 & - & -& -& - & 1.00\rpm0.00 &  1.00\rpm0.00 & 26.70\rpm25.45 & 1.00\rpm0.00 & - & -& -& -\\ 
& & \cmark & 1.00\rpm0.00 &  1.00\rpm0.00 & 5.90\rpm0.57 & 1.00\rpm0.00 & - & -& -& - & 1.00\rpm0.00 &  1.00\rpm0.00 & 36.30\rpm32.09 & \textbf{1.00\rpm0.01} & - & -& -& - \\ 
\cmidrule{2-19}
&HDDDM$_{I}$& -  & 1.00\rpm0.00 &  1.00\rpm0.00 & 10.20\rpm5.69 & 1.00\rpm0.00 & - & -& -& - & 1.00\rpm0.00 &  1.00\rpm0.00 & 38.50\rpm16.28 & 1.00\rpm0.00 & - & -& -& -\\ 
\bottomrule

\end{tabular*}
\end{sidewaystable*}

\begin{figure}[t]
    \centering
    \includegraphics[width=0.75\linewidth]{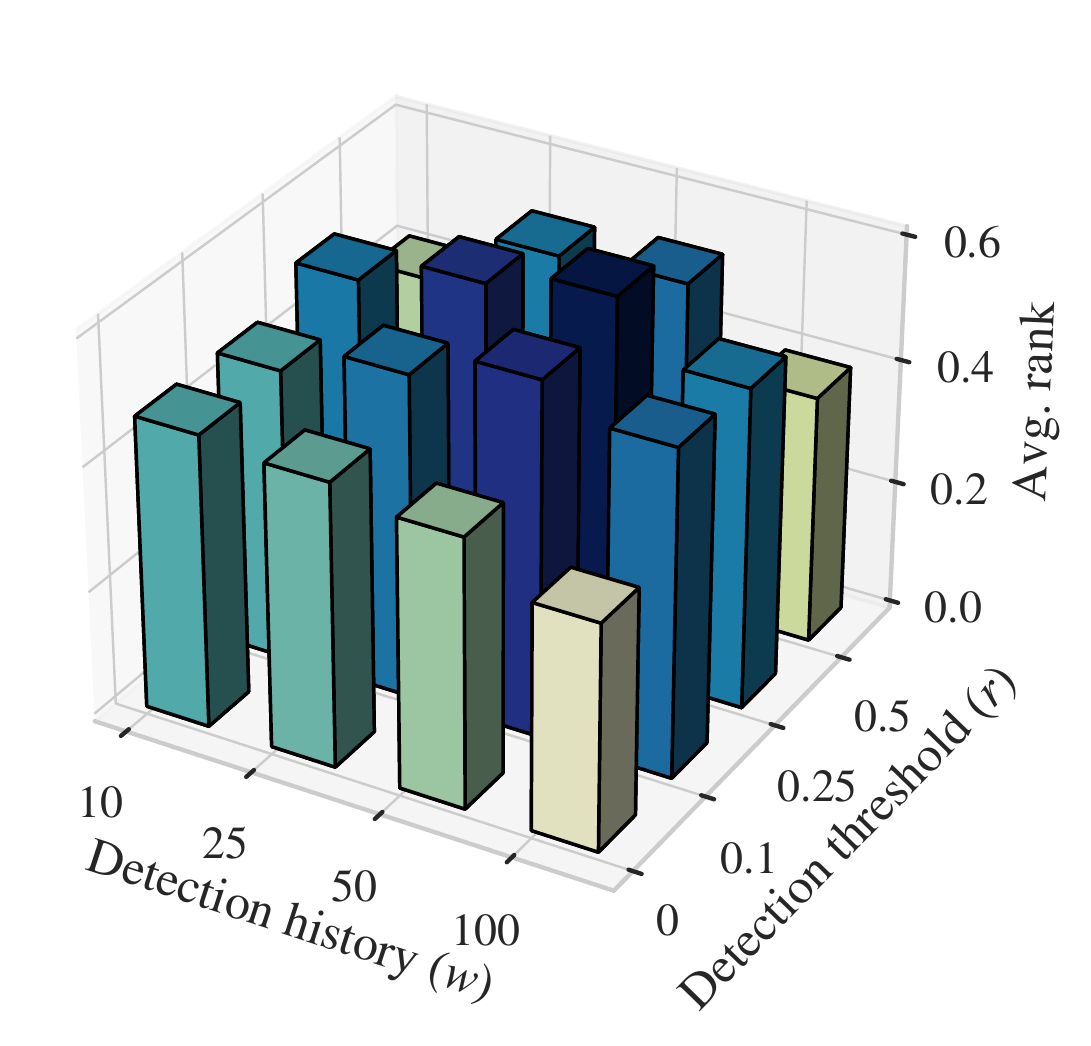}
    \caption{Average performance rank when varying drift window size $w$ and drift detection threshold $r$ calculated for each dataset and algorithm combination (higher is better).}
    \label{fig:w_sens3D}
\end{figure}

Finally, we report ablation studies on the hyperparameters introduced with the proposed framework, in particular the drift detection history size $w$ and the detection threshold ratio $r$. 
In Fig.~\ref{fig:w_sens3D}, we report the average rank of the selected hyperparameter combination $(w,r)$ spanning a grid with $w \in \{ 10, 25, 50, 100 \} $ and $r \in \{0, 0.1, 0.25, 0.50 \}$.
For very small $r\approx 0$ a drift is reported as soon as a single changed data sample is detected, which leads to many false positives for small $r$ and the corresponding low performance. 
With a large detection history and detection threshold, i.e.\ ($w=100$ and $r=0.5$), the performance are also low, because of the large delay for the detection.
The best results are achieved with a detection window of 50 samples and a detection threshold ratio from $0.1$ to $0.5$.

\section{Conclusion and future work}
In this work, we 
presented a novel semi-supervised task-sensitive concept drift detection framework, which is based on learning a constrained low-dimensional embedding representation.
The constraint imposed on the low dimensional embedding during the learning of the classification task leads to well separated clusters for each class, and allows for a robust detection of drift in incoming novel data samples. 
Any unsupervised change detection algorithm can be used for analyzing the statistics of the distances between incoming data samples and the centroids in the embedding representation and thereby detecting drifted individual samples. 
We also introduced two simple detection methods, \acrshort{ema} and \acrshort{zscore}, which are  based on the exponential moving average  and on a modified $z$-score, respectively. 
To compare the performance of different drift detection algorithms, we introduced a bounded and continuous metric $H$, which accumulates the detection accuracy, delay and false positive rate into one value.

We conducted experiments on nine benchmarks datasets where different types of drift have been induced.
The results indicated that the proposed framework is capable of robustly detecting real concept drifts, i.e.\ when the changes in the data have an impact on the classification results, while properly ignoring virtual drift, where the classification accuracy is not affected by the drift.
Furthermore, the proposed change detectors \acrshort{ema} and \acrshort{zscore} have shown  performance comparable with other SotA algorithms.
We thoroughly analyzed the dependence of the two introduced hyperparameters and identified a range of values for robust and good operation.

In future work, we target an extension of the framework to online learning settings by including active learning methods, where the embedding representation is adjusted in cases of successfully detected drifts. 

\bibliographystyle{IEEEtran}
\bibliography{biblio}

\comment{
\clearpage
\section{Supplementary material}
In this section, we report the supplementary material and additional results.
In Table~\ref{tab:GV} is reported the Generalized Variance (GV) \cite{gvwilks1932} of the class-clusters in the embedding space, with (\xmark) and without (\cmark) the presence of the constraint embedding module.
In the last column, is reported the percentage of variance reduction due to the constraining module. 

In Fig.~\ref{fig:embedding viualization}, is visualized the embedding space without (a) and with (b) the constrained module for the dataset \textit{MovRBF}. 
It is clearly observable the effect of the drift, which moves the class centroids from the initial to the final location, indicated with the arrow.
Notice how more compactly the data is distributed around the centroids, due the effect of the constraints.

In Table~\ref{tab:results_step} and Table~\ref{tab:results_grad} are respectively reported the results with step and gradual corruption of all the dataset.
Recall that the detection delay is reported only in the experiments with real drift. 
While, with virtual drift, the goal is to not detect anything.

\begin{table}[tbh]
    \centering
    \setlength{\tabcolsep}{3pt}
    \renewcommand{\arraystretch}{1}
    \aboverulesep = 0.605mm 
    \belowrulesep = 0.984mm 
    
    \caption{Generalized Variance (GV) of the class clusters in the embedding space, with (\cmark) and without (\xmark) constraints. 
    In the last column, the percentage of variance reduction due the constraint.
    }
    \begin{tabular*}{\linewidth}{l @{\extracolsep{\fill}} c c c}
\toprule
\textbf{Dataset} & $GV_{\text{\xmark}}$ & $GV_{\text{\cmark}}$ &  $\big( 1 - \frac{GV_{\text{\cmark}}}{GV_{\text{\xmark}}} \big) _\%$\\
\midrule
\textit{adult} & 3.85e-05\rpm3.46e-05 & 4.42e-07\rpm2.34e-07 & 98.85\rpm1.64 \\
\textit{bank} & 6.71e-06\rpm5.83e-06 & 3.06e-07\rpm2.19e-07 & 95.44\rpm7.24 \\
\textit{digits08} & 4.18e-06\rpm6.75e-06 & 1.54e-06\rpm1.44e-06 & 63.31\rpm40.43 \\
\textit{digits17} & 7.66e-06\rpm1.87e-06 & 1.01e-06\rpm8.07e-07 & 86.86\rpm13.74 \\
\textit{musk} & 2.87e-05\rpm3.14e-05 & 3.15e-06\rpm3.78e-06 & 89.01\rpm25.23 \\
\textit{phishing} & 7.79e-05\rpm5.58e-05 & 4.34e-06\rpm4.19e-06 & 94.43\rpm9.37 \\
\textit{wine} & 8.78e-06\rpm8.42e-06 & 2.97e-08\rpm3.35e-08 & 99.66\rpm0.71 \\
\midrule
\textit{RBF} & 6.55e-02\rpm1.55e-01 & 5.53e-03\rpm1.33e-02 & 91.56\rpm40.27 \\
\textit{MovRBF}  & 1.58e-05\rpm3.08e-05 & 8.10e-09\rpm4.23e-08 & 99.95\rpm0.37 \\
\bottomrule
    \end{tabular*}
    \label{tab:GV}
\end{table}

\begin{figure}[tbh!]
    \centering
    \includegraphics[width=0.8\linewidth]{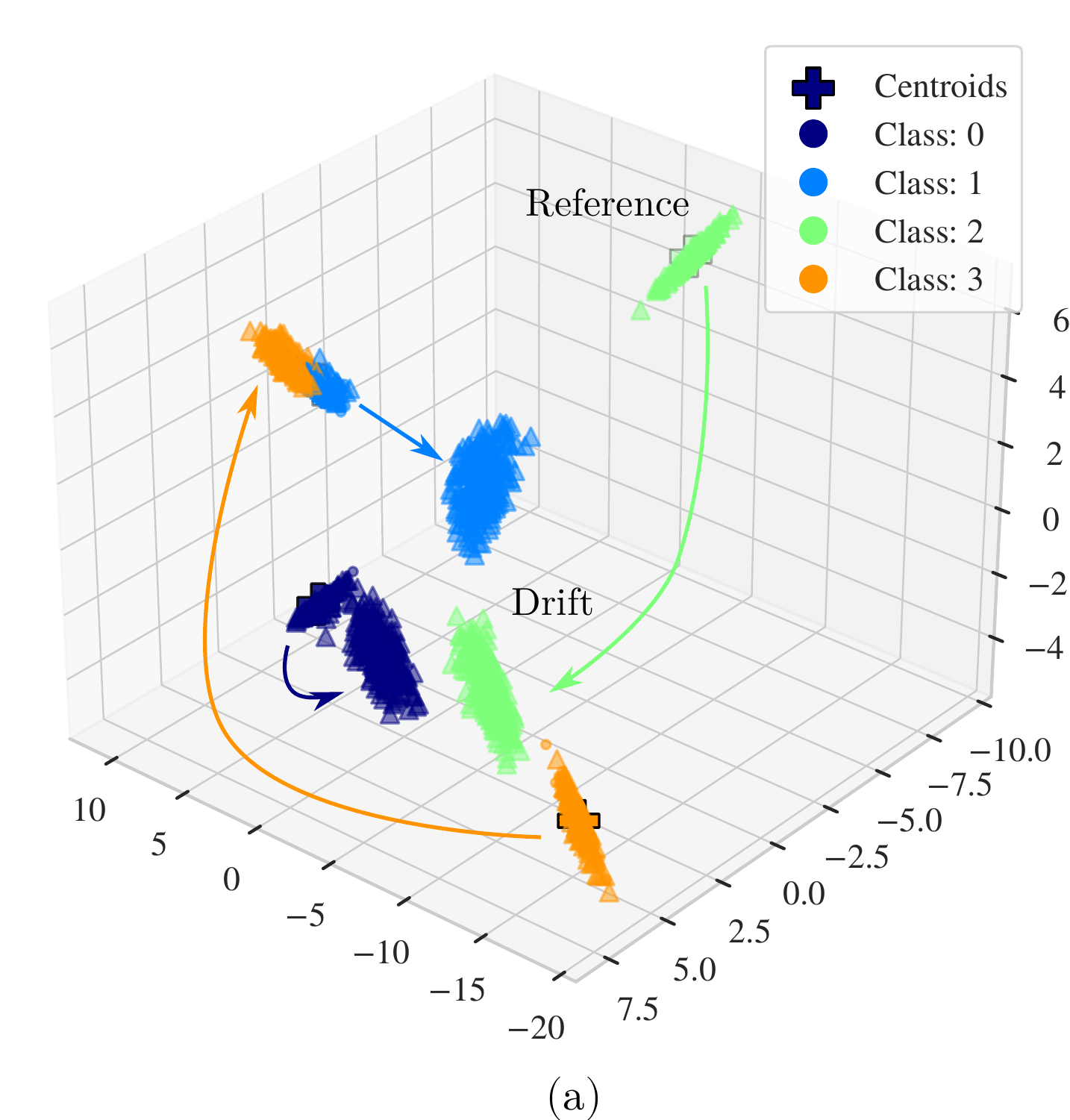}
    \includegraphics[width=0.8\linewidth]{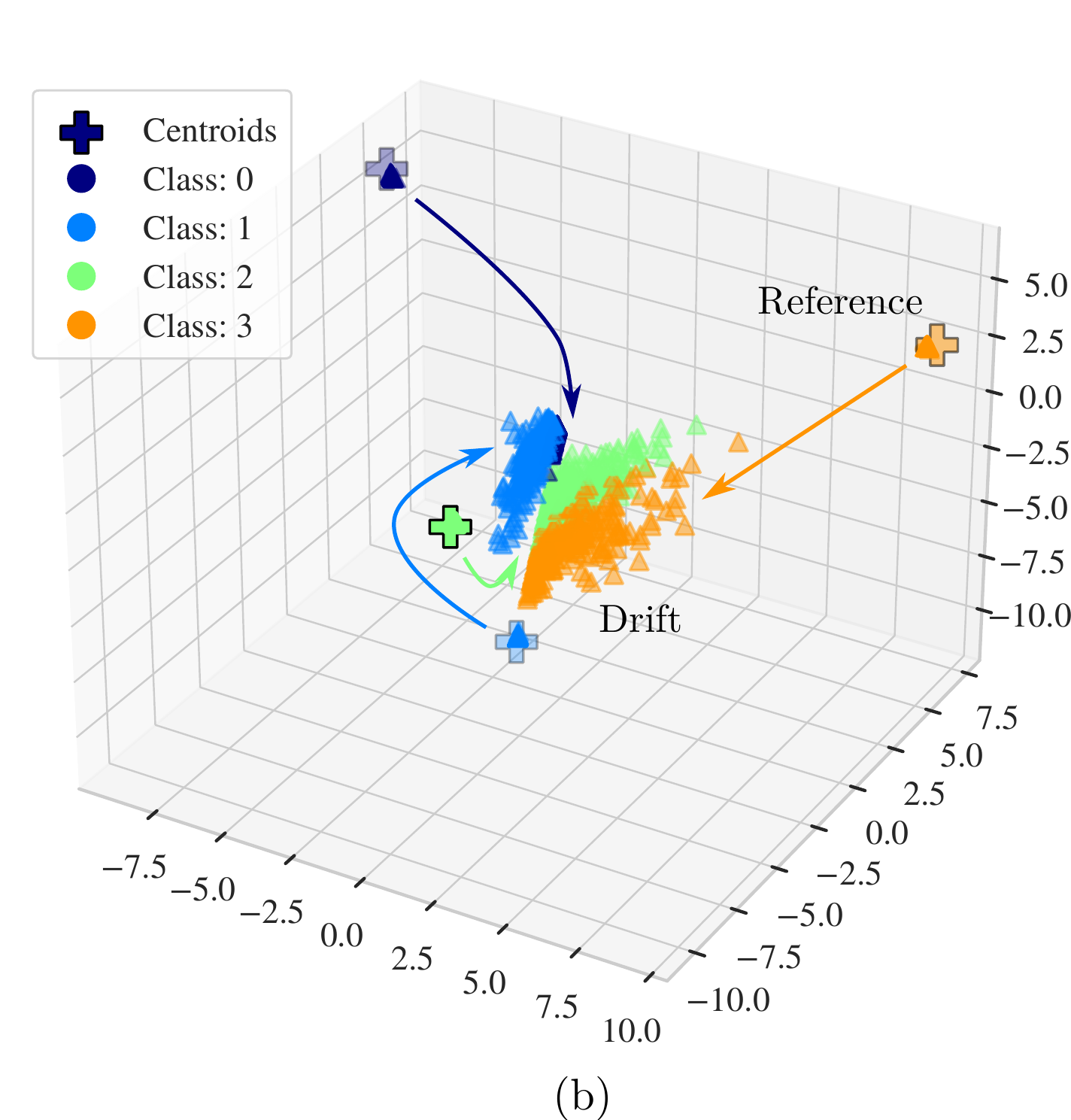}
    \caption{Visualization of the embedding of \textit{MovingRBF} dataset before and after the step drift. 
    Top (a) without, bottom (b) with the proposed constraint module.
    Notice how the reference data is more compactly distributed around the centroids with the constraint module.}
    \label{fig:embedding viualization}
\end{figure}

\begin{table*}[tb!]
    \centering
    \setlength{\tabcolsep}{4pt}
    \renewcommand{\arraystretch}{0.5}
    \aboverulesep = 0.305mm 
    \belowrulesep = 0.405mm 
    \caption{Step drift detection results. The presence (or not) of the constraint module is reported by $\cmark$ (or $\xmark$).}
    \label{tab:results_step}
    \begin{tabular*}{\linewidth}{l @{\extracolsep{\fill}} c  c c c c c | c c c c}
\toprule
\multirow{2}{*}{\textbf{Dataset}} & \multirow{2}{*}{\textbf{Detector}} & \multirow{2}{*}{$\xmark$/$\cmark$} & \multicolumn{4}{c |}{\textbf{Most informative}} & \multicolumn{4}{c}{\textbf{Least informative}} \\
& & & \textit{DA} & \textit{TNR} & \textit{Delay} & \textit{H} & \textit{DA} & \textit{TNR} & \textit{Delay} & \textit{H}\\
\midrule
\multirow{12}{*}{\textit{RBF}} & \multirow{2}{*}{ZSD} & \xmark & 0.90\rpm0.32 &  0.06\rpm0.04 & 0.00\rpm0.00 & 0.10\rpm0.08  & 0.10\rpm0.32 &  0.08\rpm0.06 & - & 0.00\rpm0.01 \\ 
& & \cmark & 1.00\rpm0.00 &  0.51\rpm0.27 & 17.40\rpm38.19 & 0.64\rpm0.21  & 0.40\rpm0.52 &  0.56\rpm0.30 & - & 0.31\rpm0.42 \\ 
\cmidrule{2-11}
& \multirow{2}{*}{EMAD} & \xmark & 0.10\rpm0.32 &  1.00\rpm0.00 & 0.00\rpm0.00 & 0.10\rpm0.32  & 0.90\rpm0.32 &  1.00\rpm0.00 & - & 0.90\rpm0.32 \\ 
& & \cmark & 0.90\rpm0.32 &  1.00\rpm0.00 & 90.20\rpm87.50 & 0.89\rpm0.31  & 0.80\rpm0.42 &  1.00\rpm0.00 & - & 0.80\rpm0.42 \\ 
\cmidrule{2-11}
& \multirow{2}{*}{IKS} & \xmark & 0.60\rpm0.52 &  1.00\rpm0.00 & 259.90\rpm128.96 & 0.47\rpm0.42  & 0.80\rpm0.42 &  1.00\rpm0.01 & - & 0.80\rpm0.42 \\ 
& & \cmark & 0.90\rpm0.32 &  0.99\rpm0.04 & 169.00\rpm120.60 & 0.83\rpm0.30  & 0.80\rpm0.42 &  0.99\rpm0.03 & - & 0.79\rpm0.42 \\ 
\cmidrule{2-11}
& \multirow{2}{*}{HDDDM$_{E}$} & \xmark & 0.90\rpm0.32 &  0.88\rpm0.07 & 100.40\rpm50.92 & 0.81\rpm0.29  & 0.80\rpm0.42 &  0.88\rpm0.05 & - & 0.68\rpm0.41 \\ 
& & \cmark & 1.00\rpm0.00 &  0.87\rpm0.08 & 82.20\rpm24.98 & 0.92\rpm0.05  & 0.50\rpm0.53 &  0.90\rpm0.10 & - & 0.47\rpm0.49 \\ 
\cmidrule{2-11}
&HDDDM$_{I}$& -  & 0.90\rpm0.32 &  0.89\rpm0.32 & 165.10\rpm122.68 & 0.75\rpm0.30  & 0.70\rpm0.48 &  0.89\rpm0.08 & - & 0.65\rpm0.45 \\ 
\midrule
\multirow{12}{*}{\textit{adult}} & \multirow{2}{*}{ZSD} & \xmark & 0.20\rpm0.42 &  0.81\rpm0.20 & 558.60\rpm1756.28 & 0.18\rpm0.38  & 0.80\rpm0.42 &  0.88\rpm0.08 & - & 0.74\rpm0.39 \\ 
& & \cmark & 0.90\rpm0.32 &  0.22\rpm0.06 & 128.50\rpm406.35 & 0.32\rpm0.14  & 0.00\rpm0.00 &  0.21\rpm0.04 & - & 0.00\rpm0.00 \\ 
\cmidrule{2-11}
& \multirow{2}{*}{EMAD} & \xmark & 0.00\rpm0.00 &  1.00\rpm0.00 & 0.00\rpm0.00 & 0.00\rpm0.00  & 0.20\rpm0.42 &  1.00\rpm0.00 & - & 0.20\rpm0.42 \\ 
& & \cmark & 0.00\rpm0.00 &  1.00\rpm0.00 & 0.00\rpm0.00 & 0.00\rpm0.00  & 1.00\rpm0.00 &  1.00\rpm0.00 & - & 1.00\rpm0.00 \\ 
\cmidrule{2-11}
& \multirow{2}{*}{IKS} & \xmark & 0.80\rpm0.42 &  0.85\rpm0.04 & 148.40\rpm123.00 & 0.66\rpm0.40  & 0.80\rpm0.42 &  0.80\rpm0.04 & - & 0.67\rpm0.36 \\ 
& & \cmark & 1.00\rpm0.00 &  0.69\rpm0.07 & 77.80\rpm42.49 & 0.80\rpm0.04  & 0.00\rpm0.00 &  0.74\rpm0.10 & - & 0.00\rpm0.00 \\ 
\cmidrule{2-11}
& \multirow{2}{*}{HDDDM$_{E}$} & \xmark & 1.00\rpm0.00 &  0.91\rpm0.01 & 107.60\rpm37.92 & 0.92\rpm0.03  & 0.30\rpm0.48 &  0.90\rpm0.02 & - & 0.23\rpm0.40 \\ 
& & \cmark & 1.00\rpm0.00 &  0.92\rpm0.02 & 94.60\rpm39.04 & 0.94\rpm0.03  & 0.80\rpm0.42 &  0.91\rpm0.02 & - & 0.76\rpm0.40 \\ 
\cmidrule{2-11}
&HDDDM$_{I}$& -  & 1.00\rpm0.00 &  0.91\rpm0.00 & 5.00\rpm0.00 & 0.95\rpm0.00  & 0.80\rpm0.42 &  0.91\rpm0.00 & - & 0.76\rpm0.40 \\ 
\midrule
\multirow{12}{*}{\textit{bank}} & \multirow{2}{*}{ZSD} & \xmark & 0.10\rpm0.32 &  0.93\rpm0.04 & 2.50\rpm7.91 & 0.09\rpm0.30  & 0.80\rpm0.42 &  0.93\rpm0.03 & - & 0.78\rpm0.41 \\ 
& & \cmark & 1.00\rpm0.00 &  0.15\rpm0.04 & 9.20\rpm14.99 & 0.26\rpm0.06  & 0.00\rpm0.00 &  0.16\rpm0.04 & - & 0.00\rpm0.00 \\ 
\cmidrule{2-11}
& \multirow{2}{*}{EMAD} & \xmark & 0.00\rpm0.00 &  1.00\rpm0.00 & 190.50\rpm602.41 & 0.00\rpm0.00  & 1.00\rpm0.00 &  1.00\rpm0.00 & - & 1.00\rpm0.00 \\ 
& & \cmark & 0.60\rpm0.52 &  1.00\rpm0.00 & 578.40\rpm1171.43 & 0.54\rpm0.47  & 1.00\rpm0.00 &  1.00\rpm0.00 & - & 1.00\rpm0.00 \\ 
\cmidrule{2-11}
& \multirow{2}{*}{IKS} & \xmark & 0.80\rpm0.42 &  0.97\rpm0.02 & 228.80\rpm158.23 & 0.68\rpm0.39  & 1.00\rpm0.00 &  0.96\rpm0.02 & - & 0.98\rpm0.01 \\ 
& & \cmark & 0.90\rpm0.32 &  0.98\rpm0.02 & 84.40\rpm75.87 & 0.82\rpm0.33  & 1.00\rpm0.00 &  0.94\rpm0.06 & - & 0.97\rpm0.03 \\ 
\cmidrule{2-11}
& \multirow{2}{*}{HDDDM$_{E}$} & \xmark & 1.00\rpm0.00 &  0.94\rpm0.03 & 111.30\rpm42.91 & 0.93\rpm0.05  & 0.70\rpm0.48 &  0.93\rpm0.02 & - & 0.67\rpm0.46 \\ 
& & \cmark & 0.90\rpm0.32 &  0.92\rpm0.03 & 141.00\rpm88.64 & 0.78\rpm0.32  & 0.30\rpm0.48 &  0.92\rpm0.03 & - & 0.29\rpm0.47 \\ 
\cmidrule{2-11}
&HDDDM$_{I}$& -  & 1.00\rpm0.00 &  0.95\rpm0.00 & 10.00\rpm1.41 & 0.98\rpm0.00  & 0.00\rpm0.00 &  0.95\rpm0.00 & - & 0.00\rpm0.00 \\ 
\midrule
\multirow{12}{*}{\textit{digits08}} & \multirow{2}{*}{ZSD} & \xmark & 1.00\rpm0.00 &  0.94\rpm0.14 & 11.80\rpm16.34 & 0.96\rpm0.09  & 0.50\rpm0.53 &  0.99\rpm0.04 & 18.50\rpm44.91 & 0.49\rpm0.51 \\ 
& & \cmark & 0.90\rpm0.32 &  1.00\rpm0.00 & 7.10\rpm3.96 & 0.90\rpm0.32  & 1.00\rpm0.00 &  1.00\rpm0.00 & 17.30\rpm5.54 & 1.00\rpm0.00 \\ 
\cmidrule{2-11}
& \multirow{2}{*}{EMAD} & \xmark & 0.90\rpm0.32 &  1.00\rpm0.00 & 23.00\rpm38.27 & 0.90\rpm0.32  & 0.50\rpm0.53 &  1.00\rpm0.00 & 11.30\rpm35.73 & 0.50\rpm0.52 \\ 
& & \cmark & 0.90\rpm0.32 &  1.00\rpm0.00 & 9.50\rpm3.54 & 0.90\rpm0.32  & 1.00\rpm0.00 &  1.00\rpm0.00 & 26.10\rpm27.91 & 1.00\rpm0.01 \\ 
\cmidrule{2-11}
& \multirow{2}{*}{IKS} & \xmark & 0.60\rpm0.52 &  1.00\rpm0.00 & 92.10\rpm60.28 & 0.60\rpm0.51  & 0.60\rpm0.52 &  1.00\rpm0.00 & 72.50\rpm97.90 & 0.57\rpm0.49 \\ 
& & \cmark & 0.90\rpm0.32 &  1.00\rpm0.00 & 41.20\rpm9.89 & 0.90\rpm0.32  & 1.00\rpm0.00 &  1.00\rpm0.00 & 73.20\rpm9.65 & 0.99\rpm0.00 \\ 
\cmidrule{2-11}
& \multirow{2}{*}{HDDDM$_{E}$} & \xmark & 0.50\rpm0.53 &  1.00\rpm0.00 & 145.00\rpm0.00 & 0.47\rpm0.50  & 0.70\rpm0.48 &  1.00\rpm0.00 & 146.70\rpm59.24 & 0.63\rpm0.44 \\ 
& & \cmark & 0.80\rpm0.42 &  1.00\rpm0.00 & 130.60\rpm45.89 & 0.75\rpm0.40  & 1.00\rpm0.00 &  1.00\rpm0.00 & 145.20\rpm0.63 & 0.94\rpm0.00 \\ 
\cmidrule{2-11}
&HDDDM$_{I}$& -  & 0.50\rpm0.53 &  1.00\rpm0.53 & 158.00\rpm21.28 & 0.45\rpm0.48  & 0.40\rpm0.52 &  1.00\rpm0.00 & 0.00\rpm0.00 & 0.40\rpm0.52 \\ 
\midrule
\multirow{12}{*}{\textit{digits17}} & \multirow{2}{*}{ZSD} & \xmark & 0.70\rpm0.48 &  0.99\rpm0.03 & 0.00\rpm0.00 & 0.70\rpm0.48  & 0.60\rpm0.52 &  0.98\rpm0.06 & - & 0.59\rpm0.51 \\ 
& & \cmark & 0.80\rpm0.42 &  0.95\rpm0.12 & 25.80\rpm28.82 & 0.77\rpm0.41  & 1.00\rpm0.00 &  0.91\rpm0.15 & - & 0.94\rpm0.09 \\ 
\cmidrule{2-11}
& \multirow{2}{*}{EMAD} & \xmark & 0.70\rpm0.48 &  1.00\rpm0.00 & 0.00\rpm0.00 & 0.70\rpm0.48  & 0.20\rpm0.42 &  1.00\rpm0.00 & - & 0.20\rpm0.42 \\ 
& & \cmark & 0.80\rpm0.42 &  1.00\rpm0.00 & 68.30\rpm79.61 & 0.76\rpm0.41  & 1.00\rpm0.00 &  1.00\rpm0.00 & - & 0.92\rpm0.07 \\ 
\cmidrule{2-11}
& \multirow{2}{*}{IKS} & \xmark & 0.50\rpm0.53 &  1.00\rpm0.00 & 46.70\rpm98.63 & 0.50\rpm0.53  & 0.20\rpm0.42 &  1.00\rpm0.00 & - & 0.20\rpm0.42 \\ 
& & \cmark & 1.00\rpm0.00 &  1.00\rpm0.00 & 145.60\rpm23.71 & 0.93\rpm0.04  & 0.90\rpm0.32 &  1.00\rpm0.00 & - & 0.81\rpm0.29 \\ 
\cmidrule{2-11}
& \multirow{2}{*}{HDDDM$_{E}$} & \xmark & 0.50\rpm0.53 &  1.00\rpm0.00 & 118.40\rpm64.81 & 0.48\rpm0.51  & 0.60\rpm0.52 &  1.00\rpm0.00 & - & 0.51\rpm0.44 \\ 
& & \cmark & 1.00\rpm0.00 &  1.00\rpm0.00 & 150.70\rpm28.88 & 0.92\rpm0.07  & 1.00\rpm0.00 &  1.00\rpm0.00 & - & 0.93\rpm0.03 \\ 
\cmidrule{2-11}
&HDDDM$_{I}$& -  & 0.30\rpm0.48 &  1.00\rpm0.48 & 133.20\rpm2.10 & 0.29\rpm0.46  & 0.80\rpm0.42 &  1.00\rpm0.00 & - & 0.66\rpm0.35 \\ 
\midrule
\multirow{12}{*}{\textit{musk}} & \multirow{2}{*}{ZSD} & \xmark & 0.20\rpm0.42 &  0.78\rpm0.19 & 241.90\rpm372.12 & 0.15\rpm0.32  & 0.70\rpm0.48 &  0.83\rpm0.13 & - & 0.64\rpm0.45 \\ 
& & \cmark & 1.00\rpm0.00 &  0.99\rpm0.02 & 33.60\rpm10.79 & 0.99\rpm0.01  & 0.10\rpm0.32 &  0.98\rpm0.02 & - & 0.10\rpm0.31 \\ 
\cmidrule{2-11}
& \multirow{2}{*}{EMAD} & \xmark & 0.00\rpm0.00 &  1.00\rpm0.00 & 0.00\rpm0.00 & 0.00\rpm0.00  & 0.80\rpm0.42 &  1.00\rpm0.00 & - & 0.80\rpm0.42 \\ 
& & \cmark & 1.00\rpm0.00 &  1.00\rpm0.00 & 53.20\rpm29.12 & 0.99\rpm0.01  & 0.40\rpm0.52 &  1.00\rpm0.00 & - & 0.38\rpm0.49 \\ 
\cmidrule{2-11}
& \multirow{2}{*}{IKS} & \xmark & 0.20\rpm0.42 &  1.00\rpm0.00 & 318.10\rpm370.49 & 0.13\rpm0.28  & 0.80\rpm0.42 &  1.00\rpm0.00 & - & 0.80\rpm0.42 \\ 
& & \cmark & 1.00\rpm0.00 &  1.00\rpm0.01 & 59.60\rpm11.36 & 0.99\rpm0.01  & 0.10\rpm0.32 &  1.00\rpm0.00 & - & 0.09\rpm0.30 \\ 
\cmidrule{2-11}
& \multirow{2}{*}{HDDDM$_{E}$} & \xmark & 1.00\rpm0.00 &  0.87\rpm0.07 & 121.80\rpm62.01 & 0.87\rpm0.06  & 0.30\rpm0.48 &  0.86\rpm0.06 &-  & 0.25\rpm0.40 \\ 
& & \cmark & 1.00\rpm0.00 &  0.95\rpm0.08 & 27.40\rpm14.29 & 0.97\rpm0.04  & 0.10\rpm0.32 &  0.96\rpm0.04 & - & 0.10\rpm0.31 \\ 
\cmidrule{2-11}
&HDDDM$_{I}$& -  & 1.00\rpm0.00 &  0.96\rpm0.00 & 0.00\rpm0.00 & 0.98\rpm0.00  & 0.20\rpm0.42 &  0.96\rpm0.00 &  & 0.20\rpm0.41 \\ 
\midrule
\multirow{12}{*}{\textit{phishing}} & \multirow{2}{*}{ZSD} & \xmark & 0.90\rpm0.32 &  0.47\rpm0.21 & 80.80\rpm155.72 & 0.55\rpm0.28  & 0.50\rpm0.53 &  0.44\rpm0.29 & 261.10\rpm331.56 & 0.30\rpm0.38 \\ 
& & \cmark & 1.00\rpm0.00 &  0.08\rpm0.04 & 0.00\rpm0.00 & 0.14\rpm0.07  & 0.70\rpm0.48 &  0.13\rpm0.07 & 0.00\rpm0.00 & 0.19\rpm0.16 \\ 
\cmidrule{2-11}
& \multirow{2}{*}{EMAD} & \xmark & 0.10\rpm0.32 &  1.00\rpm0.00 & 6.30\rpm19.92 & 0.10\rpm0.31  & 0.70\rpm0.48 &  1.00\rpm0.00 & 0.00\rpm0.00 & 0.70\rpm0.48 \\ 
& & \cmark & 0.70\rpm0.48 &  1.00\rpm0.00 & 197.50\rpm214.28 & 0.66\rpm0.46  & 0.30\rpm0.48 &  1.00\rpm0.00 & 0.00\rpm0.00 & 0.30\rpm0.48 \\ 
\cmidrule{2-11}
& \multirow{2}{*}{IKS} & \xmark & 1.00\rpm0.00 &  0.71\rpm0.09 & 9.20\rpm15.96 & 0.83\rpm0.06  & 0.30\rpm0.48 &  0.66\rpm0.05 & 0.00\rpm0.00 & 0.24\rpm0.38 \\ 
& & \cmark & 1.00\rpm0.00 &  0.78\rpm0.12 & 16.60\rpm35.15 & 0.87\rpm0.08  & 0.60\rpm0.52 &  0.73\rpm0.14 & 44.80\rpm119.99 & 0.51\rpm0.44 \\ 
\cmidrule{2-11}
& \multirow{2}{*}{HDDDM$_{E}$} & \xmark & 1.00\rpm0.00 &  0.82\rpm0.04 & 111.10\rpm39.41 & 0.87\rpm0.05  & 0.70\rpm0.48 &  0.81\rpm0.03 & 403.30\rpm606.53 & 0.63\rpm0.43 \\ 
& & \cmark & 1.00\rpm0.00 &  0.78\rpm0.03 & 132.40\rpm42.85 & 0.83\rpm0.06  & 0.40\rpm0.52 &  0.77\rpm0.03 & 146.20\rpm386.87 & 0.32\rpm0.42 \\ 
\cmidrule{2-11}
&HDDDM$_{I}$& -  & 1.00\rpm0.00 &  0.84\rpm0.00 & 14.30\rpm2.50 & 0.91\rpm0.00  & 0.30\rpm0.48 &  0.84\rpm0.00 & 17.70\rpm4.00 & 0.27\rpm0.44 \\ 
\midrule
\multirow{12}{*}{\textit{wine}} & \multirow{2}{*}{ZSD} & \xmark & 1.00\rpm0.00 &  0.67\rpm0.13 & 19.30\rpm21.97 & 0.80\rpm0.10  & 0.60\rpm0.52 &  0.72\rpm0.19 & - & 0.55\rpm0.47 \\ 
& & \cmark & 1.00\rpm0.00 &  1.00\rpm0.00 & 28.30\rpm13.15 & 1.00\rpm0.00  & 1.00\rpm0.00 &  1.00\rpm0.00 & - & 1.00\rpm0.00 \\ 
\cmidrule{2-11}
& \multirow{2}{*}{EMAD} & \xmark & 0.00\rpm0.00 &  1.00\rpm0.00 & 0.00\rpm0.00 & 0.00\rpm0.00  & 1.00\rpm0.00 &  1.00\rpm0.00 & - & 1.00\rpm0.00 \\ 
& & \cmark & 1.00\rpm0.00 &  0.95\rpm0.02 & 13.50\rpm5.64 & 0.97\rpm0.01  & 1.00\rpm0.00 &  0.95\rpm0.02 & - & 0.98\rpm0.01 \\ 
\cmidrule{2-11}
& \multirow{2}{*}{IKS} & \xmark & 0.80\rpm0.42 &  1.00\rpm0.00 & 224.00\rpm127.86 & 0.68\rpm0.38  & 1.00\rpm0.00 &  1.00\rpm0.00 & - & 1.00\rpm0.00 \\ 
& & \cmark & 1.00\rpm0.00 &  1.00\rpm0.00 & 94.20\rpm18.60 & 0.98\rpm0.01  & 0.90\rpm0.32 &  1.00\rpm0.00 & - & 0.90\rpm0.32 \\ 
\cmidrule{2-11}
& \multirow{2}{*}{HDDDM$_{E}$} & \xmark & 1.00\rpm0.00 &  0.88\rpm0.05 & 39.60\rpm16.85 & 0.93\rpm0.03  & 0.50\rpm0.53 &  0.88\rpm0.06 & - & 0.47\rpm0.50 \\ 
& & \cmark & 1.00\rpm0.00 &  0.89\rpm0.08 & 23.40\rpm13.75 & 0.94\rpm0.05  & 0.70\rpm0.48 &  0.90\rpm0.06 & - & 0.66\rpm0.46 \\ 
\cmidrule{2-11}
&HDDDM$_{I}$& -  & 1.00\rpm0.00 &  0.88\rpm0.00 & 22.50\rpm2.42 & 0.93\rpm0.00  & 0.00\rpm0.00 &  0.88\rpm0.00 & - & 0.00\rpm0.00 \\ 

\midrule
\multirow{12}{*}{\textit{MovRBF}} & \multirow{2}{*}{ZSD} & \xmark   & 1.00\rpm0.00 &  0.82\rpm0.18 & 2.50\rpm2.07 & 0.89\rpm0.11 & - & -& -& - \\ 
& & \cmark   & 1.00\rpm0.00 &  1.00\rpm0.00 & 4.10\rpm0.32 & 1.00\rpm0.00 & - & -& -& - \\ 
\cmidrule{2-11}
& \multirow{2}{*}{EMAD} & \xmark   & 1.00\rpm0.00 &  1.00\rpm0.00 & 6.20\rpm1.75 & 1.00\rpm0.00 & - & -& -& - \\ 
& & \cmark   & 1.00\rpm0.00 &  1.00\rpm0.00 & 4.00\rpm0.00 & 1.00\rpm0.00 & - & -& -& - \\ 
\cmidrule{2-11}
& \multirow{2}{*}{IKS} & \xmark   & 1.00\rpm0.00 &  1.00\rpm0.00 & 57.20\rpm34.52 & 0.99\rpm0.01 & - & -& -& - \\ 
& & \cmark   & 1.00\rpm0.00 &  0.99\rpm0.04 & 29.30\rpm15.17 & 0.99\rpm0.02 & - & -& -& - \\ 
\cmidrule{2-11}
& \multirow{2}{*}{HDDDM$_{E}$} & \xmark   & 1.00\rpm0.00 &  1.00\rpm0.00 & 7.40\rpm1.84 & 1.00\rpm0.00 & - & -& -& - \\ 
& & \cmark   & 1.00\rpm0.00 &  1.00\rpm0.00 & 5.90\rpm0.57 & 1.00\rpm0.00 & - & -& -& - \\ 
\cmidrule{2-11}
&HDDDM$_{I}$& -    & 1.00\rpm0.00 &  1.00\rpm0.00 & 10.20\rpm5.69 & 1.00\rpm0.00 & - & -& -& - \\ 
\bottomrule

\end{tabular*}
\end{table*}

\begin{table*}[tb!]
    \centering
    \setlength{\tabcolsep}{4pt}
    \renewcommand{\arraystretch}{0.5}
    \aboverulesep = 0.305mm 
    \belowrulesep = 0.405mm 
    \caption{Gradual drift detection results. The presence (or not) of the constraint module is reported by $\cmark$ (or $\xmark$).}
    \label{tab:results_grad}
    \begin{tabular*}{\linewidth}{l @{\extracolsep{\fill}} c  c c c c c | c c c c}
\toprule
\multirow{2}{*}{\textbf{Dataset}} & \multirow{2}{*}{\textbf{Detector}} & \multirow{2}{*}{$\xmark$/$\cmark$} & \multicolumn{4}{c |}{\textbf{Most informative}} & \multicolumn{4}{c}{\textbf{Least informative}} \\
& & & DA & TNR & Delay & H & DA & TNR & Delay \\
\midrule
\multirow{12}{*}{\textit{RBF}} & \multirow{2}{*}{ZSD} & \xmark & 1.00\rpm0.00 &  0.10\rpm0.10 & 3.80\rpm9.81 & 0.17\rpm0.16  & 0.10\rpm0.32 &  0.08\rpm0.08 & - & 0.01\rpm0.03 \\ 
& & \cmark & 1.00\rpm0.00 &  0.49\rpm0.25 & 4.20\rpm6.81 & 0.62\rpm0.24  & 0.20\rpm0.42 &  0.51\rpm0.24 & - & 0.14\rpm0.29 \\ 
\cmidrule{2-11}
& \multirow{2}{*}{EMAD} & \xmark & 0.60\rpm0.52 &  1.00\rpm0.00 & 190.20\rpm163.24 & 0.45\rpm0.48  & 0.90\rpm0.32 &  1.00\rpm0.00 & - & 0.90\rpm0.32 \\ 
& & \cmark & 1.00\rpm0.00 &  1.00\rpm0.00 & 47.00\rpm48.84 & 0.99\rpm0.02  & 0.90\rpm0.32 &  1.00\rpm0.00 & - & 0.90\rpm0.32 \\ 
\cmidrule{2-11}
& \multirow{2}{*}{IKS} & \xmark & 0.90\rpm0.32 &  1.00\rpm0.00 & 207.70\rpm142.06 & 0.74\rpm0.37  & 0.90\rpm0.32 &  0.99\rpm0.04 & - & 0.89\rpm0.31 \\ 
& & \cmark & 1.00\rpm0.00 &  1.00\rpm0.00 & 85.40\rpm33.56 & 0.98\rpm0.02  & 0.50\rpm0.53 &  1.00\rpm0.01 & - & 0.50\rpm0.53 \\ 
\cmidrule{2-11}
& \multirow{2}{*}{HDDDM$_{E}$} & \xmark & 1.00\rpm0.00 &  0.88\rpm0.09 & 66.10\rpm34.42 & 0.93\rpm0.05  & 0.60\rpm0.52 &  0.89\rpm0.08 & - & 0.57\rpm0.49 \\ 
& & \cmark & 1.00\rpm0.00 &  0.91\rpm0.07 & 50.10\rpm25.91 & 0.95\rpm0.04  & 0.50\rpm0.53 &  0.93\rpm0.09 & - & 0.46\rpm0.49 \\ 
\cmidrule{2-11}
&HDDDM$_{I}$& -  & 1.00\rpm0.00 &  0.89\rpm0.00 & 41.80\rpm30.26 & 0.94\rpm0.04  & 0.10\rpm0.32 &  0.89\rpm0.08 & - & 0.10\rpm0.31 \\ 
\midrule
\multirow{12}{*}{\textit{adult}} & \multirow{2}{*}{ZSD} & \xmark & 0.00\rpm0.00 &  0.82\rpm0.13 & 1865.30\rpm1325.69 & 0.00\rpm0.00  & 0.50\rpm0.53 &  0.87\rpm0.12 & - & 0.47\rpm0.50 \\ 
& & \cmark & 1.00\rpm0.00 &  0.24\rpm0.06 & 8.60\rpm27.20 & 0.39\rpm0.08  & 0.00\rpm0.00 &  0.25\rpm0.05 & - & 0.00\rpm0.00 \\ 
\cmidrule{2-11}
& \multirow{2}{*}{EMAD} & \xmark & 0.00\rpm0.00 &  1.00\rpm0.00 & 812.40\rpm1712.89 & 0.00\rpm0.00  & 0.60\rpm0.52 &  1.00\rpm0.00 & - & 0.60\rpm0.52 \\ 
& & \cmark & 0.00\rpm0.00 &  1.00\rpm0.00 & 0.00\rpm0.00 & 0.00\rpm0.00  & 1.00\rpm0.00 &  1.00\rpm0.00 & - & 1.00\rpm0.00 \\ 
\cmidrule{2-11}
& \multirow{2}{*}{IKS} & \xmark & 1.00\rpm0.00 &  0.81\rpm0.07 & 162.30\rpm57.49 & 0.80\rpm0.02  & 0.40\rpm0.52 &  0.80\rpm0.06 & - & 0.36\rpm0.47 \\ 
& & \cmark & 1.00\rpm0.00 &  0.71\rpm0.11 & 34.70\rpm19.93 & 0.82\rpm0.07  & 0.00\rpm0.00 &  0.69\rpm0.08 & - & 0.00\rpm0.00 \\ 
\cmidrule{2-11}
& \multirow{2}{*}{HDDDM$_{E}$} & \xmark & 1.00\rpm0.00 &  0.91\rpm0.02 & 46.70\rpm10.77 & 0.95\rpm0.01  & 0.40\rpm0.52 &  0.91\rpm0.02 & - & 0.25\rpm0.40 \\ 
& & \cmark & 1.00\rpm0.00 &  0.92\rpm0.03 & 52.20\rpm10.92 & 0.96\rpm0.01  & 0.00\rpm0.00 &  0.91\rpm0.01 & - & 0.00\rpm0.00 \\ 
\cmidrule{2-11}
&HDDDM$_{I}$& -  & 1.00\rpm0.00 &  0.91\rpm0.00 & 5.00\rpm0.00 & 0.95\rpm0.00  & 0.40\rpm0.52 &  0.91\rpm0.00 & - & 0.38\rpm0.49 \\ 
\midrule
\multirow{12}{*}{\textit{bank}} & \multirow{2}{*}{ZSD} & \xmark & 0.50\rpm0.53 &  0.90\rpm0.04 & 772.10\rpm754.77 & 0.27\rpm0.41  & 0.70\rpm0.48 &  0.92\rpm0.03 & - & 0.65\rpm0.45 \\ 
& & \cmark & 1.00\rpm0.00 &  0.15\rpm0.06 & 2.60\rpm8.22 & 0.25\rpm0.09  & 0.00\rpm0.00 &  0.13\rpm0.05 & - & 0.00\rpm0.00 \\ 
\cmidrule{2-11}
& \multirow{2}{*}{EMAD} & \xmark & 0.00\rpm0.00 &  1.00\rpm0.00 & 3219.40\rpm1427.96 & 0.00\rpm0.00  & 0.70\rpm0.48 &  1.00\rpm0.00 & - & 0.70\rpm0.48 \\ 
& & \cmark & 0.30\rpm0.48 &  1.00\rpm0.00 & 179.50\rpm245.01 & 0.27\rpm0.45  & 1.00\rpm0.00 &  1.00\rpm0.00 & - & 1.00\rpm0.00 \\ 
\cmidrule{2-11}
& \multirow{2}{*}{IKS} & \xmark & 0.00\rpm0.00 &  0.96\rpm0.04 & 1050.10\rpm887.76 & 0.00\rpm0.00  & 0.60\rpm0.52 &  0.96\rpm0.02 & - & 0.59\rpm0.51 \\ 
& & \cmark & 1.00\rpm0.00 &  0.95\rpm0.03 & 56.10\rpm17.12 & 0.97\rpm0.01  & 1.00\rpm0.00 &  0.98\rpm0.01 & - & 0.99\rpm0.01 \\ 
\cmidrule{2-11}
& \multirow{2}{*}{HDDDM$_{E}$} & \xmark & 1.00\rpm0.00 &  0.93\rpm0.02 & 89.40\rpm19.78 & 0.95\rpm0.01  & 0.60\rpm0.52 &  0.94\rpm0.03 & - & 0.57\rpm0.49 \\ 
& & \cmark & 1.00\rpm0.00 &  0.94\rpm0.02 & 76.40\rpm47.22 & 0.95\rpm0.02  & 0.50\rpm0.53 &  0.93\rpm0.02 & - & 0.48\rpm0.51 \\ 
\cmidrule{2-11}
&HDDDM$_{I}$& -  & 1.00\rpm0.00 &  0.95\rpm0.00 & 6.00\rpm0.00 & 0.98\rpm0.00  & 0.30\rpm0.48 &  0.95\rpm0.00 & - & 0.29\rpm0.47 \\ 
\midrule
\multirow{12}{*}{\textit{digits08}} & \multirow{2}{*}{ZSD} & \xmark & 0.60\rpm0.52 &  0.99\rpm0.04 & 14.20\rpm25.53 & 0.60\rpm0.52  & 0.60\rpm0.52 &  1.00\rpm0.00 & - & 0.60\rpm0.52 \\ 
& & \cmark & 0.60\rpm0.52 &  1.00\rpm0.00 & 9.10\rpm2.38 & 0.60\rpm0.52  & 0.30\rpm0.48 &  1.00\rpm0.00 & - & 0.30\rpm0.48 \\ 
\cmidrule{2-11}
& \multirow{2}{*}{EMAD} & \xmark & 0.50\rpm0.53 &  1.00\rpm0.00 & 12.20\rpm29.35 & 0.50\rpm0.53  & 0.60\rpm0.52 &  1.00\rpm0.00 & - & 0.60\rpm0.52 \\ 
& & \cmark & 0.60\rpm0.52 &  1.00\rpm0.00 & 12.00\rpm2.62 & 0.60\rpm0.52  & 0.40\rpm0.52 &  1.00\rpm0.00 & - & 0.39\rpm0.50 \\ 
\cmidrule{2-11}
& \multirow{2}{*}{IKS} & \xmark & 0.70\rpm0.48 &  1.00\rpm0.00 & 108.30\rpm96.02 & 0.64\rpm0.45  & 0.70\rpm0.48 &  1.00\rpm0.00 & - & 0.68\rpm0.47 \\ 
& & \cmark & 0.60\rpm0.52 &  1.00\rpm0.00 & 76.50\rpm7.92 & 0.60\rpm0.51  & 0.30\rpm0.48 &  1.00\rpm0.00 & - & 0.30\rpm0.48 \\ 
\cmidrule{2-11}
& \multirow{2}{*}{HDDDM$_{E}$} & \xmark & 0.40\rpm0.52 &  1.00\rpm0.00 & 131.00\rpm46.06 & 0.38\rpm0.48  & 0.50\rpm0.53 &  1.00\rpm0.00 & - & 0.48\rpm0.50 \\ 
& & \cmark & 0.50\rpm0.53 &  1.00\rpm0.00 & 145.90\rpm59.64 & 0.45\rpm0.48  & 0.30\rpm0.48 &  1.00\rpm0.00 & - & 0.28\rpm0.45 \\ 
\cmidrule{2-11}
&HDDDM$_{I}$& -  & 0.50\rpm0.53 &  1.00\rpm0.53 & 0.00\rpm0.00 & 0.50\rpm0.53  & 0.60\rpm0.52 &  1.00\rpm0.00 & - & 0.60\rpm0.52 \\ 
\midrule
\multirow{12}{*}{\textit{digits17}} & \multirow{2}{*}{ZSD} & \xmark & 0.60\rpm0.52 &  0.96\rpm0.13 & 11.70\rpm18.31 & 0.57\rpm0.50  & 0.70\rpm0.48 &  0.99\rpm0.03 & - & 0.69\rpm0.48 \\ 
& & \cmark & 1.00\rpm0.00 &  0.99\rpm0.04 & 4.00\rpm0.00 & 0.99\rpm0.02  & 1.00\rpm0.00 &  0.99\rpm0.04 & - & 0.95\rpm0.07 \\ 
\cmidrule{2-11}
& \multirow{2}{*}{EMAD} & \xmark & 0.40\rpm0.52 &  1.00\rpm0.00 & 24.50\rpm25.62 & 0.40\rpm0.52  & 0.60\rpm0.52 &  1.00\rpm0.00 & - & 0.60\rpm0.52 \\ 
& & \cmark & 1.00\rpm0.00 &  1.00\rpm0.00 & 7.80\rpm0.42 & 1.00\rpm0.00  & 0.40\rpm0.52 &  1.00\rpm0.00 & - & 0.34\rpm0.44 \\ 
\cmidrule{2-11}
& \multirow{2}{*}{IKS} & \xmark & 0.40\rpm0.52 &  1.00\rpm0.00 & 92.80\rpm16.85 & 0.40\rpm0.51  & 0.60\rpm0.52 &  1.00\rpm0.00 & - & 0.60\rpm0.52 \\ 
& & \cmark & 1.00\rpm0.00 &  1.00\rpm0.00 & 57.50\rpm2.59 & 1.00\rpm0.00  & 1.00\rpm0.00 &  1.00\rpm0.00 & - & 0.92\rpm0.08 \\ 
\cmidrule{2-11}
& \multirow{2}{*}{HDDDM$_{E}$} & \xmark & 0.40\rpm0.52 &  1.00\rpm0.00 & 132.00\rpm0.00 & 0.38\rpm0.49  & 0.50\rpm0.53 &  1.00\rpm0.00 & - & 0.48\rpm0.51 \\ 
& & \cmark & 0.90\rpm0.32 &  1.00\rpm0.00 & 141.10\rpm66.63 & 0.80\rpm0.31  & 0.90\rpm0.32 &  1.00\rpm0.00 & - & 0.78\rpm0.29 \\ 
\cmidrule{2-11}
&HDDDM$_{I}$& -  & 0.40\rpm0.52 &  1.00\rpm0.52 & 132.00\rpm0.00 & 0.38\rpm0.49  & 0.40\rpm0.52 &  1.00\rpm0.00 & - & 0.38\rpm0.49 \\ 
\midrule
\multirow{12}{*}{\textit{musk}} & \multirow{2}{*}{ZSD} & \xmark & 1.00\rpm0.00 &  0.87\rpm0.15 & 11.60\rpm2.63 & 0.92\rpm0.09  & 1.00\rpm0.00 &  0.80\rpm0.13 & - & 0.88\rpm0.08 \\ 
& & \cmark & 1.00\rpm0.00 &  0.97\rpm0.03 & 8.20\rpm2.04 & 0.99\rpm0.01  & 1.00\rpm0.00 &  0.98\rpm0.03 & - & 0.99\rpm0.01 \\ 
\cmidrule{2-11}
& \multirow{2}{*}{EMAD} & \xmark & 1.00\rpm0.00 &  1.00\rpm0.00 & 21.60\rpm3.72 & 1.00\rpm0.00  & 0.70\rpm0.48 &  1.00\rpm0.00 & - & 0.63\rpm0.45 \\ 
& & \cmark & 1.00\rpm0.00 &  1.00\rpm0.00 & 14.10\rpm2.47 & 1.00\rpm0.00  & 1.00\rpm0.00 &  1.00\rpm0.00 & - & 0.99\rpm0.01 \\ 
\cmidrule{2-11}
& \multirow{2}{*}{IKS} & \xmark & 1.00\rpm0.00 &  1.00\rpm0.00 & 70.10\rpm13.08 & 0.99\rpm0.00  & 1.00\rpm0.00 &  1.00\rpm0.01 & - & 0.95\rpm0.03 \\ 
& & \cmark & 1.00\rpm0.00 &  1.00\rpm0.00 & 47.60\rpm9.02 & 1.00\rpm0.00  & 1.00\rpm0.00 &  1.00\rpm0.00 & - & 0.99\rpm0.01 \\ 
\cmidrule{2-11}
& \multirow{2}{*}{HDDDM$_{E}$} & \xmark & 1.00\rpm0.00 &  0.88\rpm0.06 & 25.50\rpm8.09 & 0.93\rpm0.03  & 1.00\rpm0.00 &  0.88\rpm0.06 & - & 0.93\rpm0.03 \\ 
& & \cmark & 1.00\rpm0.00 &  0.95\rpm0.04 & 16.20\rpm7.27 & 0.97\rpm0.02  & 1.00\rpm0.00 &  0.93\rpm0.07 & - & 0.96\rpm0.04 \\ 
\cmidrule{2-11}
&HDDDM$_{I}$& -  & 1.00\rpm0.00 &  0.96\rpm0.00 & 0.00\rpm0.00 & 0.98\rpm0.00  & 1.00\rpm0.00 &  0.96\rpm0.00 & - & 0.98\rpm0.00 \\ 
\midrule
\multirow{12}{*}{\textit{phishing}} & \multirow{2}{*}{ZSD} & \xmark & 0.70\rpm0.48 &  0.39\rpm0.17 & 201.80\rpm203.94 & 0.34\rpm0.26  & 0.40\rpm0.52 &  0.58\rpm0.20 & 273.30\rpm377.38 & 0.33\rpm0.43 \\ 
& & \cmark & 1.00\rpm0.00 &  0.13\rpm0.07 & 0.00\rpm0.00 & 0.22\rpm0.11  & 0.90\rpm0.32 &  0.13\rpm0.06 & 0.00\rpm0.00 & 0.22\rpm0.12 \\ 
\cmidrule{2-11}
& \multirow{2}{*}{EMAD} & \xmark & 0.20\rpm0.42 &  1.00\rpm0.00 & 0.00\rpm0.00 & 0.20\rpm0.42  & 0.90\rpm0.32 &  1.00\rpm0.00 & 0.00\rpm0.00 & 0.90\rpm0.32 \\ 
& & \cmark & 0.00\rpm0.00 &  1.00\rpm0.00 & 0.00\rpm0.00 & 0.00\rpm0.00  & 0.10\rpm0.32 &  1.00\rpm0.00 & 0.00\rpm0.00 & 0.10\rpm0.32 \\ 
\cmidrule{2-11}
& \multirow{2}{*}{IKS} & \xmark & 0.70\rpm0.48 &  0.68\rpm0.09 & 0.00\rpm0.00 & 0.55\rpm0.38  & 0.20\rpm0.42 &  0.70\rpm0.07 & 0.00\rpm0.00 & 0.17\rpm0.36 \\ 
& & \cmark & 0.70\rpm0.48 &  0.73\rpm0.17 & 249.40\rpm353.25 & 0.50\rpm0.36  & 0.70\rpm0.48 &  0.76\rpm0.08 & 186.20\rpm438.38 & 0.59\rpm0.41 \\ 
\cmidrule{2-11}
& \multirow{2}{*}{HDDDM$_{E}$} & \xmark & 0.30\rpm0.48 &  0.78\rpm0.04 & 558.20\rpm402.23 & 0.23\rpm0.38  & 0.90\rpm0.32 &  0.81\rpm0.06 & 164.80\rpm408.80 & 0.80\rpm0.28 \\ 
& & \cmark & 0.20\rpm0.42 &  0.79\rpm0.03 & 443.30\rpm349.13 & 0.16\rpm0.34  & 0.20\rpm0.42 &  0.75\rpm0.04 & 369.90\rpm593.49 & 0.17\rpm0.35 \\ 
\cmidrule{2-11}
&HDDDM$_{I}$& -  & 0.80\rpm0.42 &  0.84\rpm0.42 & 12.00\rpm0.00 & 0.73\rpm0.38  & 0.10\rpm0.32 &  0.84\rpm0.00 & 11.00\rpm0.00 & 0.09\rpm0.29 \\ 
\midrule
\multirow{12}{*}{\textit{wine}} & \multirow{2}{*}{ZSD} & \xmark & 1.00\rpm0.00 &  0.79\rpm0.10 & 18.50\rpm12.26 & 0.88\rpm0.06  & 0.10\rpm0.32 &  0.63\rpm0.18 & - & 0.09\rpm0.30 \\ 
& & \cmark & 1.00\rpm0.00 &  1.00\rpm0.00 & 99.80\rpm11.43 & 0.98\rpm0.01  & 1.00\rpm0.00 &  1.00\rpm0.00 & - & 1.00\rpm0.00 \\ 
\cmidrule{2-11}
& \multirow{2}{*}{EMAD} & \xmark & 1.00\rpm0.00 &  1.00\rpm0.00 & 68.30\rpm64.54 & 0.96\rpm0.12  & 1.00\rpm0.00 &  1.00\rpm0.00 & - & 1.00\rpm0.00 \\ 
& & \cmark & 1.00\rpm0.00 &  0.95\rpm0.02 & 41.30\rpm28.06 & 0.97\rpm0.02  & 1.00\rpm0.00 &  0.95\rpm0.03 & - & 0.97\rpm0.01 \\ 
\cmidrule{2-11}
& \multirow{2}{*}{IKS} & \xmark & 1.00\rpm0.00 &  1.00\rpm0.00 & 58.40\rpm8.82 & 1.00\rpm0.00  & 1.00\rpm0.00 &  1.00\rpm0.00 & - & 1.00\rpm0.00 \\ 
& & \cmark & 1.00\rpm0.00 &  1.00\rpm0.00 & 61.70\rpm25.49 & 0.99\rpm0.01  & 0.30\rpm0.48 &  1.00\rpm0.00 & - & 0.30\rpm0.48 \\ 
\cmidrule{2-11}
& \multirow{2}{*}{HDDDM$_{E}$} & \xmark & 1.00\rpm0.00 &  0.88\rpm0.04 & 42.60\rpm9.99 & 0.93\rpm0.02  & 0.30\rpm0.48 &  0.92\rpm0.06 & - & 0.29\rpm0.47 \\ 
& & \cmark & 1.00\rpm0.00 &  0.85\rpm0.05 & 37.80\rpm11.35 & 0.92\rpm0.03  & 0.40\rpm0.52 &  0.89\rpm0.06 & - & 0.37\rpm0.48 \\ 
\cmidrule{2-11}
&HDDDM$_{I}$& -  & 1.00\rpm0.00 &  0.88\rpm0.00 & 44.90\rpm1.97 & 0.93\rpm0.00  & 0.00\rpm0.00 &  0.88\rpm0.00 & - & 0.00\rpm0.00 \\ 

\midrule
\multirow{12}{*}{\textit{MovRBF}} & \multirow{2}{*}{ZSD} & \xmark & 1.00\rpm0.00 &  0.86\rpm0.00 & 5.80\rpm4.05 & 0.89\rpm0.22 & - & -& -& - \\ 
& & \cmark & 1.00\rpm0.00 &  1.00\rpm0.00 & 65.40\rpm40.07 & 0.99\rpm0.01 & - & -& -& - \\ 
\cmidrule{2-11}
& \multirow{2}{*}{EMAD} & \xmark & 1.00\rpm0.00 &  1.00\rpm0.00 & 23.10\rpm28.78 & 1.00\rpm0.01 & - & -& -& - \\ 
& & \cmark & 1.00\rpm0.00 &  1.00\rpm0.00 & 17.70\rpm9.87 & 1.00\rpm0.00 & - & -& -& - \\ 
\cmidrule{2-11}
& \multirow{2}{*}{IKS} & \xmark & 1.00\rpm0.00 &  1.00\rpm0.00 & 77.70\rpm30.74 & 0.99\rpm0.02 & - & -& -& - \\ 
& & \cmark & 1.00\rpm0.00 &  1.00\rpm0.00 & 50.80\rpm22.46 & 1.00\rpm0.00 & - & -& -& - \\ 
\cmidrule{2-11}
& \multirow{2}{*}{HDDDM$_{E}$} & \xmark & 1.00\rpm0.00 &  1.00\rpm0.00 & 26.70\rpm25.45 & 1.00\rpm0.00 & - & -& -& - \\ 
& & \cmark & 1.00\rpm0.00 &  1.00\rpm0.00 & 36.30\rpm32.09 & 1.00\rpm0.01 & - & -& -& - \\ 
\cmidrule{2-11}
&HDDDM$_{I}$& -  & 1.00\rpm0.00 &  1.00\rpm0.00 & 38.50\rpm16.28 & 1.00\rpm0.00 & - & -& -& - \\ 
\bottomrule

\end{tabular*}
\end{table*}

}

\end{document}